\pdfoutput=1
\documentclass[11pt]{article}

\usepackage[]{acl}

\usepackage{times}
\usepackage{latexsym}
\usepackage{subfiles}
\usepackage{graphicx}
\usepackage{booktabs}
\usepackage{diagbox}
\usepackage{slashbox}
\usepackage{float}
\usepackage{textcomp}
\usepackage{enumitem}
\usepackage{xcolor}
\usepackage{colortbl}
\usepackage{todonotes}
\usepackage{multirow}
\usepackage{subcaption}
\usepackage{amsmath}
\usepackage{dsfont}
\usepackage{algorithm}
\usepackage{algorithmic}
\usepackage[T1]{fontenc}
\usepackage[utf8]{inputenc}

\usepackage{microtype}

\newcommand{\telin}[1]{{\color{blue}{\small\bf\sf [Te-Lin: #1]}}}

\newcommand{\Skip}[1]{}

\newcommand{\ie}{\textit{i}.\textit{e}.\ }
\newcommand{\eg}{\textit{e}.\textit{g}.\ }
\newcommand{\Eg}{\textit{E}.\textit{g}.\ }

\newcommand{\secref}[1]{Section \ref{#1}}
\newcommand{\figref}[1]{Figure \ref{#1}}
\newcommand{\tbref}[1]{Table \ref{#1}}

\newcommand{\appendixref}[1]{Append. Sec. \ref{#1}}

\newcommand{\dotieconcat}[2]{% auxiliary macro, don't use it directly
  \text{\raisebox{.8ex}{$\smallfrown$}}%
}

\newcommand{\mypar}[1]{\noindent\textbf{#1}}

\newcommand{\SideNote}[2]{\todo[color=#1,size=\small]{#2}} 
\newcommand{\violet}[1]{\SideNote{purple!40}{#1 --Violet}}

\title{Understanding Multimodal Procedural Knowledge by Sequencing Multimodal Instructional Manuals}

{\centering
    \author{
        Te-Lin Wu$^{1}$,
        Alex Spangher$^{2}$,
        Pegah Alipoormolabashi$^3$,
        \\
        \textbf{
        Marjorie Freedman$^2$,
        Ralph Weischedel$^2$,
        Nanyun Peng$^1$
        }
        \\
    $^1$University of California, Los Angeles,
    $^2$ISI, University of Southern California,\\
    $^3$Sharif University of Technology\\
    \texttt{\{telinwu,violetpeng\}@cs.ucla.edu}, \texttt{palipoor976@gmail.com} \\
    \texttt{\{spangher,mrf,weisched\}@isi.edu} \\
}
}

\begin{document}

\maketitle

\begin{abstract}
    % Human knowledge about how to perform a task (\eg ``how to make a wood sign'') is often communicated through a set of procedural instructions with a combination of texts and images.

\iffalse
The ability to order unordered events is evidence of common sense enabling humans to reason about tasks (the world) multimodally.
Such capability can benefit applications such as...
However, xxx remains an open question
In this work, we aim at benchmarking models as well as improve the models with our pretraining techniques
\fi

The ability to sequence unordered events is evidence of comprehension and reasoning about real world tasks/procedures. It is essential for applications such as task planning and multi-source instruction summarization.
It often requires thorough understanding of temporal common sense and multimodal information, since these procedures are often conveyed by a combination of texts and images.
While humans are capable of reasoning about and sequencing unordered procedural instructions, %as well as utilizing multimodality to facilitate such a task, 
%it remains a challenge for current machine learning models.
the extent to which the current machine learning methods possess such capability is still an open question.
In this work, we benchmark models' capability of reasoning over and sequencing \textit{unordered multimodal} instructions by curating datasets from online instructional manuals and collecting comprehensive human annotations.
We find current state-of-the-art models not only perform significantly worse than humans but also seem incapable of efficiently utilizing  multimodal information.
% \violet{add some conclusion here: e.g., machines are far behind human and seem to be less efficient than human on utilizing multimodal information.} 
To improve machines' performance on multimodal event sequencing, we propose sequence-aware pretraining techniques exploiting sequential alignment properties of both texts and images, resulting in \textgreater5\% improvements on perfect match ratio.
\end{abstract}

\section{Introduction}

Instructions are essential sources for agents to learn how to complete complex tasks composed of multiple steps (e.g., ``making a wood sign from scratch'').
However, %processes
instructions 
do not always come in a proper sequential order, for example,  %processes inherent in a narrative or in the news.
when instructions must be combined across sources (e.g., to accomplish a complex task there might be \textit{multiple} useful resources for certain task-steps come out from a single Google search).
%Furthermore, in situations such as multitasking, humans have developed the ability to reason about task sequences that are not in proper sequential order \cite{bode2009decoding, broeker2018multitasking}.
Therefore, \textit{sequencing \textbf{unordered} task-steps} is crucial for comprehending and inferring task procedures, which requires thorough understanding of event causal and temporal common sense.
It is essential for applications such as multi-source instruction summarization and robot task planning~\cite{garattoni2018autonomous}.
% Progress in task sequencing can help filling incomplete task instructions and robot task planning~\cite{garattoni2018autonomous}.

\begin{figure}[t]
\centering
    \includegraphics[width=1.0\columnwidth]{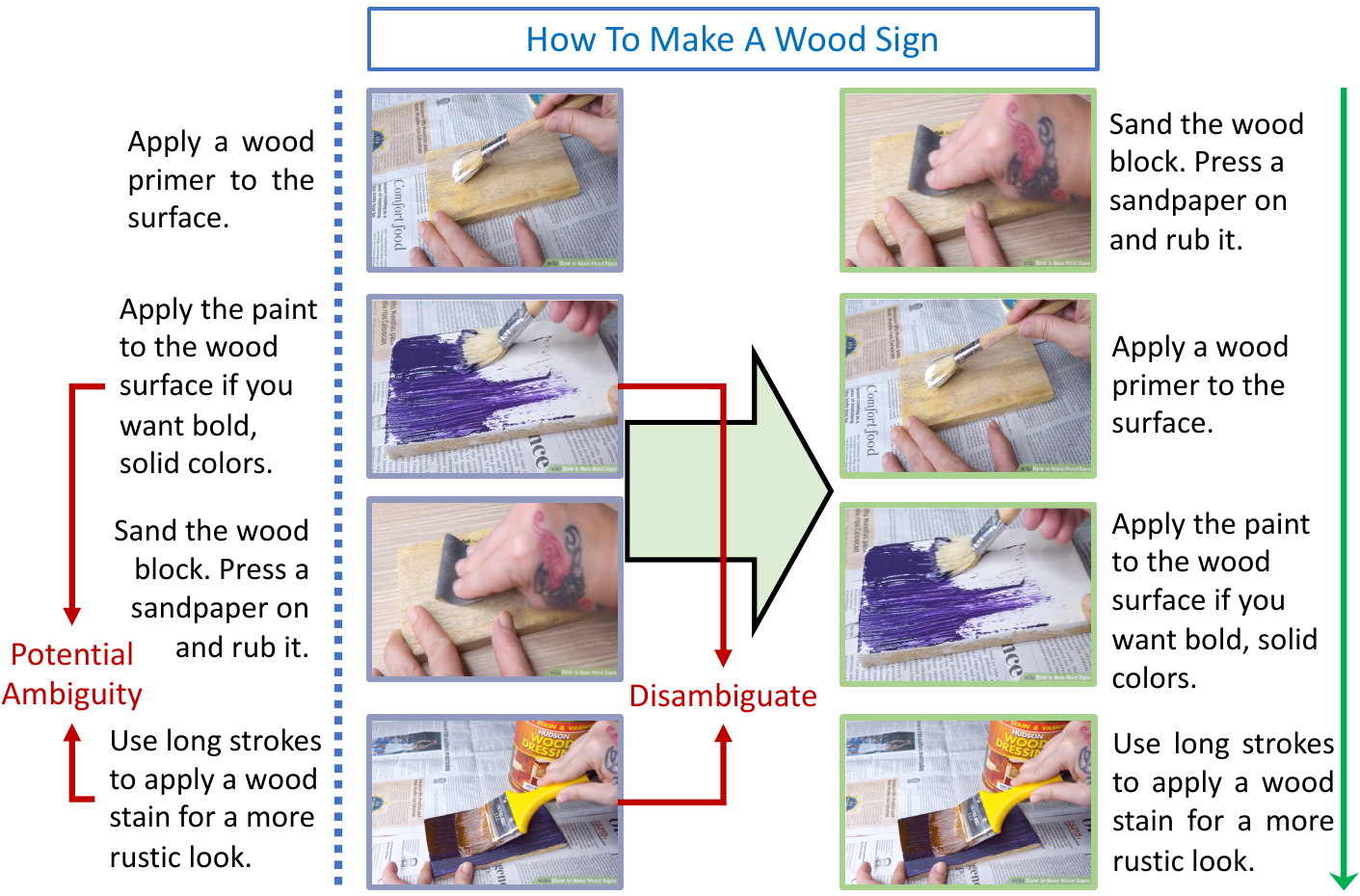}
    \caption{ \footnotesize
        \textbf{Multimodal task procedure sequencing:} The left column shows unordered instruction steps from the manual~\textit{How To Make Wood Signs}. Each step is a text description and its associated image. Without the complementary information from the visuals, a novice may have difficulty inferring the proper task order. Considering multimodal information, the proper order can be correctly inferred (right column).
    }
    \label{fig:teaser}
    \vspace{-1.em}
\end{figure}

Existing work has studied sequencing unordered \textit{texts} from paper abstracts or short stories~\cite{chen2016neural, cui2018deep}. However, real-life tasks are often complex, and multimodal information is usually provided to supplement textual descriptions to avoid ambiguity or illustrate details that are hard to narrate, as illustrated in~\figref{fig:teaser}.
\footnote{Our datasets can be accessed at \href{https://drive.google.com/drive/folders/1wQjR62nKfkh8Q77Le4H43_kAaKNFEwco?usp=sharing}{this link}.}
% Sequencing such tasks is still under explored.
% and script generation~\cite{wang2017integrating}.
% In computer vision, sequencing video frames is shown to learn useful representations~\cite{lee2017unsupervised, li2020hero}.
% \violet{I comment this out. No need to get into this in the intro -- we'll just discuss in the related work.}
%While certain prior work also attempts to extend the sequencing task to incorporate multimodality~\cite{agrawal2016sort}, the dataset used, Visual StoryTelling~\cite{huang2016visual}, merely targets \textit{short human created stories}; the featured images were not intended to be procedural (or temporally coherent) nor supply unstated details to complement the texts\footnote{The images come from photo albums and the annotators \textit{create} stories based on their freely arranged image sequences.}.
% However, the images in the employed dataset, Visual StoryTelling~\cite{huang2016visual}, were either not intended to be procedural (and temporally coherent) or created to supply unstated details to the texts\footnote{The images come from photo albums and the annotators \textit{create} stories based on their freely arranged image sequences.}.

To investigate whether current AI techniques can efficiently leverage multimodal information to sequence unordered task instructions, we curate two datasets from \textit{online instructional manuals}~\cite{wikihow, yagcioglu2018recipeqa}.
We consider two representative instruction domains: cooking recipes and ``How-To" instructions (WikiHow). 
We establish human performance for the sequencing task on a subset of each data resource.
As certain steps to perform a task can potentially be interchangeable,\footnote{\Eg, slicing carrots or cucumbers for a dish preparation, does not necessarily need to follow a specific order.} we collect annotations of possible \textbf{orders} alternative to the originally authored ones to create \textit{multiple references}. %that in reality can be performed in a different order.
Such additional annotation provides not only better measurement of human and model performance by alleviating unintended biases from content creators, but also a useful resource for future research of models that are aware of task-step dependencies and interchangeability.

% for better measuring of human and model performances. %gauge the interchangeability of certain task-steps.
% , construct a \textbf{multi-reference} dataset.

% \violet{should probably talk about interchangeability here as well. and say we collected annotation. e.g., we curate two... and collect human annotations to gauge the interchangeability xxx and establish human performance.}

To measure the ability of state-of-the-art AI techniques to sequence instruction steps, we construct models consisting of:
(1) an \textbf{input encoder} which encodes image, text, or multimodal inputs, 
and (2) an \textbf{order decoder} which predicts step order using the encoded representations.
They are  jointly trained with the order supervisions.

% We employ RoBERTa~\cite{liu2019roberta} and two strong vision-language (V\&L) models: VisualBERT~\cite{li2019visualbert} and CLIP-ViL~\cite{shen2021much}
% , mainly differ by how the visual inputs are encoded,
% , for the text-only and multimodal encoders,
% \telins{Probably can cut these even shorter?}
% and integrate a recent sentence ordering framework, BERSON~\cite{cuietal2020bert}, for the decoder.

Our preliminary studies show that multimodal information is consistently helpful for the sequencing task.
However, compared to humans, current models are less %do not seem\telins{is "seem" a good word? but I also don't want to make this too strong to prevent some random attack lol} able to 
efficient in utilizing multimodal information.
We hypothesize that it is because the models do not effectively capture the sequential information in the vision modality nor the sequential alignment between multimodal contents.
To address this, we propose to equip models with capabilities of performing \textbf{sequential aware multimodal grounding}.
%The encoders are thus first pretrained on the instruction resources through several
Specifically, we propose several self-supervised objectives, including sequence-based masked language modeling, image region modeling, and content swapped prediction, to pretrain the models before finetuning them on the downstream sequencing task.

%\telins{I omit all the details of the objectives suggested by I-Hung, do you think this is good?}
\iffalse
(1) sequence-based multimodal masked language modeling (\textbf{MLM}),
(2) image-swapping prediction (\textbf{ISP}) which asks the models if some images are not following proper order in a sequence,
(3) patch-based image-swapping prediction (\textbf{PISP}) which is similar to ISP but acts on the image-patch level,
and (4) sequence-based masked region modeling (\textbf{SMRM}) which requires the models to reconstruct masked image patches with a dynamically constructed visual vocabulary in the mini-batch training.
\fi

% \violet{I'll move this forward the merge it with the previous discussion about data collection.}

%\violet{here, I'm thinking about the high-level messages we want to send. Is it important to argue that multimodal information is helpful? why this is important? Should we instead emphasize that the current model is still quite weak in incorporating multi-modal info while humans are very good at it? \telin{I've tried to revise like the current version.}}

The proposed pretraining techniques are shown to be effective in improving multimodal performance, enjoying a \textgreater5\% improvement on the perfect match ratio metric.
However, it is still significantly behind human performance ($\sim 15$\% in perfect match ratio metric). 
The same trend is observed when alternative orders are considered.
% show that for both standard and multi-reference versions of our task, multimodal information is consistently helpful, and our proposed pretraining techniques can further improve multimodal models' performance.
% We conduct in-depth analysis to investigate different aspects of the sequencing task, specifically we incorporate the category information of WikiHow to provide a more detailed breakdown comparing human and model performance, as well as quantifying the impact when evaluating with multiple reference orders.
% We show that even when alternative orders are considered, our findings of multimodality being helpful still hold and our proposed sequencing task remains challenging.

% the interchangeability of certain task-orders along with evaluating models' performance while considering alternative orders.
%\violet{what's the take home message from your analysis? \telin{Added.}}

Our key contributions are two-fold:
(1) We propose a multimodal sequencing task with two curated instructional manuals, and comprehensive human annotations.
(2) We investigate model performance on sequencing unordered manuals, and propose sequence-aware pretraining techniques to more effectively use the multimodal information.
%is helpful on such procedure understanding task
% yet a significant gap ($\sim15$\%) well below human performance is still observed.
%(3) We conduct extensive analysis on our curated datasets as well as model performance to provide more in-depth insights.
Our experiments and extensive analysis provide insights on which task categories are most challenging for the state-of-the-art models. They also shed the light that more sophisticated sequential multimodal grounding are required to further improve the performance for the proposed multimodal sequencing task.

\section{Problem Definition}

Given a task procedure $S$ consisting of $N$ steps, where each step $S_i \in S$ can consist of two types of contents: a textual description $T_i$ of tokens $\{T_{i,k}\}_{k=1}^{n_T}$ and/or image(s) $I_i=\{I_{i,k}\}_{k=1}^{n_I}$.\footnote{For computational concerns, we set $n_I=1$ in this work.} 
A model is required to take as inputs a random permutation of $S$, \ie $S_p = \{S_{p_1}, ..., S_{p_N}\}$, where $p$ is a permutation ($S_{p_j}$ can take one of the following three modalities: $T_{p_j}$, $I_{p_j}$, and $\{T_{p_j},I_{p_j}\}$), and predict the correct order of $S_p$, \ie $\text{argsort}(S_p)$.

\section{Datasets and Human Annotation}
We are interested in understanding the current state-of-the-art models' performance on this multimodal instruction sequencing task. To this end, we curate instruction datasets to support our study.

\subsection{Instruction Manual Datasets}

There are three major features we require for the target datasets:
(1) It is multimodal.
(2) It consists of task procedures as sequences of steps. %, where different steps show \textit{procedural dependencies} in each modality.
% \telins{Counter the VIST in SortStory more?}
(3) Different modalities are used intentionally to complement each other.
In light of these, we consider the following two datasets:

\vspace{.3em}

\mypar{RecipeQA.}
% \telin{Details of the RecipeQA dataset and our transformation.}
% \telin{Quick statistics?}
We start from a popular as well as intuitive choice of instruction manuals, recipes, which fully fulfill the aforementioned criteria.
RecipeQA is a multimodal question answering dataset consisting of recipes scraped from \textit{Instructables.com}~\cite{yagcioglu2018recipeqa}.
We utilize the recipes collected in RecipeQA and convert each unique recipe into sequential multimodal steps for our task.

\vspace{.3em}

\mypar{WikiHow.}
% \telin{Details of the WikiHow dataset.}
% \telin{The scraping and cleaning procedures.}
% \telin{The hierarchy information.}
% \telin{The difference between ours and the abstractive WikiHow dataset as a contribution?}
% \telin{Quick statistics?}
To expand the types of instruction manuals for our task beyond recipes, we also consider a popular ``How To ..." type of instructions, WikiHow, which is an online knowledge base that consists of human-created articles describing procedures to accomplish a desired task.
Each article contains a high level goal of a task, a short summary of the task procedures, and several \textit{multimodal} steps where each step consists of a description paired with one or a few corresponding images.

We scrape the entire WikiHow knowledge resource, containing more than 100k unique articles (mostly) with multimodal contents
% \footnote{The dataset and relevant tools will be made public.}
, as well as the hierarchically structured category for each article.
\tbref{tab:data-det} presents the essential statistics of the two datasets
% WikiHow in~\tbref{tab:data-det-wikihow}, and RecipeQA in~\tbref{tab:data-det-recipeqa}.
(more details are in~\appendixref{a-sec:dataset_stats}).

\begin{table}[t!]
\begin{subtable}{\columnwidth}
\centering
\small
\scalebox{0.9}{
\begin{tabular}{lrrrr}
    \toprule
    \multicolumn{1}{c}{\textbf{Type}} & \multicolumn{4}{c}{\textbf{Counts}} \\
    \midrule
    \multicolumn{1}{c}{Total Unique Articles} & \multicolumn{4}{c}{109486} \\
    \multicolumn{1}{c}{Total Unique Images} & \multicolumn{4}{c}{1521909} \\
    \multicolumn{1}{c}{Train / Dev / Golden-Test} & \multicolumn{4}{c}{98268 / 11218 / 300} \\
    \multicolumn{1}{c}{Type-Token Ratio} & \multicolumn{4}{c}{216434 / 82396591 = 0.0026} \\
    \toprule
    \multicolumn{1}{c}{\textbf{Type}} & \multicolumn{1}{c}{\textbf{Mean}} & \multicolumn{1}{c}{\textbf{Std}} & \multicolumn{1}{c}{\textbf{Min}} & \multicolumn{1}{c}{\textbf{Max}} \\
    \midrule
    \multicolumn{1}{l}{Tokens in a Step Text}     & 52.95 & 26.25 & 0 & 5339 \\
    \multicolumn{1}{l}{Sentences in a Step Text}  & 3.36 & 1.3 & 0 & 50 \\
    \multicolumn{1}{l}{Number of Steps of a Task} & 5.27 & 2.62 & 0 & 75 \\
    \bottomrule
\end{tabular}
}
\caption{\footnotesize WikiHow}
\label{tab:data-det-wikihow}
\end{subtable}

\vspace{.5em}

\begin{subtable}{\columnwidth}
\centering
\small
\scalebox{0.9}{
\begin{tabular}{lrrrr}
    \toprule
    \multicolumn{1}{c}{\textbf{Type}} & \multicolumn{4}{c}{\textbf{Counts}} \\
    \midrule
    \multicolumn{1}{c}{Total Unique Articles} & \multicolumn{4}{c}{10063} \\
    \multicolumn{1}{c}{Total Unique Images} & \multicolumn{4}{c}{87840} \\
    \multicolumn{1}{c}{Train / Dev / Golden-Test} & \multicolumn{4}{c}{8032 / 2031 / 100} \\
    \multicolumn{1}{c}{Type-Token Ratio} & \multicolumn{4}{c}{91443 / 5324859 = 0.017} \\
    \toprule
    \multicolumn{1}{c}{\textbf{Type}} & \multicolumn{1}{c}{\textbf{Mean}} & \multicolumn{1}{c}{\textbf{Std}} & \multicolumn{1}{c}{\textbf{Min}} & \multicolumn{1}{c}{\textbf{Max}} \\
    \midrule
    \multicolumn{1}{l}{Tokens in a Step Text}     & 82.08 & 84.72 & 0 & 998 \\
    \multicolumn{1}{l}{Sentences in a Step Text}  & 4.19 & 4.22 & 0 & 73 \\
    \multicolumn{1}{l}{Number of Steps of a Task} & 6.45 & 2.57 & 4 & 20 \\
    \bottomrule
\end{tabular}
}
\caption{\footnotesize RecipeQA}
\label{tab:data-det-recipeqa}
\end{subtable}

\caption{\footnotesize
\textbf{General statistics of the two datasets}: We provide the detailed component counts of the datasets used in this work, including the statistics of tokens and sentences from the instruction steps (lower half of the two tables).
}
\label{tab:data-det}
\end{table}

\subsection{Human Performance Benchmark}
\label{sec:human_annots}

To ensure the validity of our proposed multimodal sequencing task, we establish the human performance via Amazon Mechanical Turk.
Since our dataset is constructed from resources that are not directly designed for the sequencing task, the quality of random samples is unverified. Specifically, some articles in WikiHow may not have a notion of proper order among the steps.\footnote{No temporal or other dependencies among the task-steps, \eg ``How to be a good person'', where each step depicts a different aspect and tips of being a good person.}
As a result, to construct a high quality test set particularly for WikiHow for establishing human performance, we first identify a set of categories which are more likely to feature proper order, \eg \textit{Home and Garden} and \textit{Hobbies and Crafts}.\footnote{Although the data used for training is not cleansed and thus can be noisy, we believe models can still learn to sequence from many of the articles designed to have proper order.} A random proportion is then sampled and the co-authors further downsample the subset to 300 samples with the aforementioned criteria via majority vote.
% : procedural in both texts and images and complementary multimodality.
For RecipeQA, we randomly sample 100 recipes from the dataset.
And hence, the resulting two subsets serve as our \textbf{golden-test-set} for performance benchmarking.

\vspace{.3em}

\mypar{Human Performance.}
Prompted with a task goal and a randomly scrambled sequence of the task-steps (can be one of the following modalities: multimodal or text/image-only), workers are asked to examine the contents and decide the proper performing order.
Human performance are then computed against the original authored orders as the ground truths, averaged across the whole set.\footnote{We design an algorithm to compute the inter-annotator agreements (IAAs), see~\appendixref{a-sec:iaa} for details. The IAAs for (\textit{multimodal}, \textit{text-only}, \textit{image-only}) versions in WikiHow is: (0.84, 0.82, 0.69), and (0.92, 0.87, 0.81) in RecipeQA.}
%\violet{I feel this section should be separated into two, one just talk about data sample, another talk about how do you collect human annotation to establish human performance. the latter is currently missing. \telin{how about now?}}

%
% In addition to labelling the correct order of the unordered manual,

% Apart from the sequencing task, we also ask the annotators for their confidence of predictions and if multimodality is helpful for deciding the order.
% For more details, see ~\appendixref{a-sec:human}.

\vspace{.3em}

\mypar{Alternative Orders.}
%In this work, we seek to examine the sequential understanding of models on instruction steps from the aforementioned manuals, however, we observe that it is not always true to enforce an only way of the step ordering. 
When performing a task, some steps can be interchangeable. %Humans are capable of understand the \textit{interchangeability}. To investigate this,
To take the interchangeability into consideration in our benchmark task, we also collect possible alternative orders to the original ones to create multiple references.
For each instance in our golden-test-set, given the instruction steps sequenced in their original order, we ask workers to annotate alternative orders if the presented task-steps can be performed following a different order.\footnote{The alternative order annotation IAAs for (\textit{multimodal}, \textit{text-only}, \textit{image-only}) versions in WikiHow is: (0.73, 0.71, 0.78), and (0.79, 0.76, 0.79) in RecipeQA.} %where we ask the workers to list possible orders alternative to the original authored ones. 
%The inter-annotator agreement is sufficiently high (see~\secref{sec:iaa} and~\secref{sec:eval_alternative_orders}), we retain the alternative orders which are majority agreed upon for each instance in our datasets.

Although in this work we are mainly focusing on sequential instructions and hence the interchangeability is also gauged in a sequential manner, we want to point out that the nature of task-step interchangeability is also highly related to parallel (branching) steps of tasks~\cite{sakaguchi2021}.
We argue that the actions that can be performed interchangeably imply no direct dependencies are among these actions and thus can potentially be parallelized, and hence our alternative order formulation can help inferring these parallel actions.

More details of the two human annotation tasks can be found in~\appendixref{a-sec:human}.

\section{Models}

% \telin{summarize our main contributions}
% The model, which takes as inputs a set of \textit{unordered} task-steps and predicts their appropriate ordering,
% considered for the proposed multimodal sequencing task
To benchmark the proposed task, we construct models comprising:
(1) an \textbf{encoder} which encodes multimodal or text/image-only inputs,
and (2) an \textbf{order decoder} which utilizes the encoded representations to predict the orders.
To help models capture \textbf{sequentiality in task-steps} better as well as adapt to our target task domains, we pretrain the encoders with several self-supervised objectives on the instructions before integrating them with the decoder.

\subsection{Input Encoders}

% The proposed sequencing task can have variants across three following input modalities: text-only, image-only, and multimodal, which correspond to the following three different encoders.

\begin{figure*}[ht!]
\centering
    \includegraphics[width=.9\textwidth]{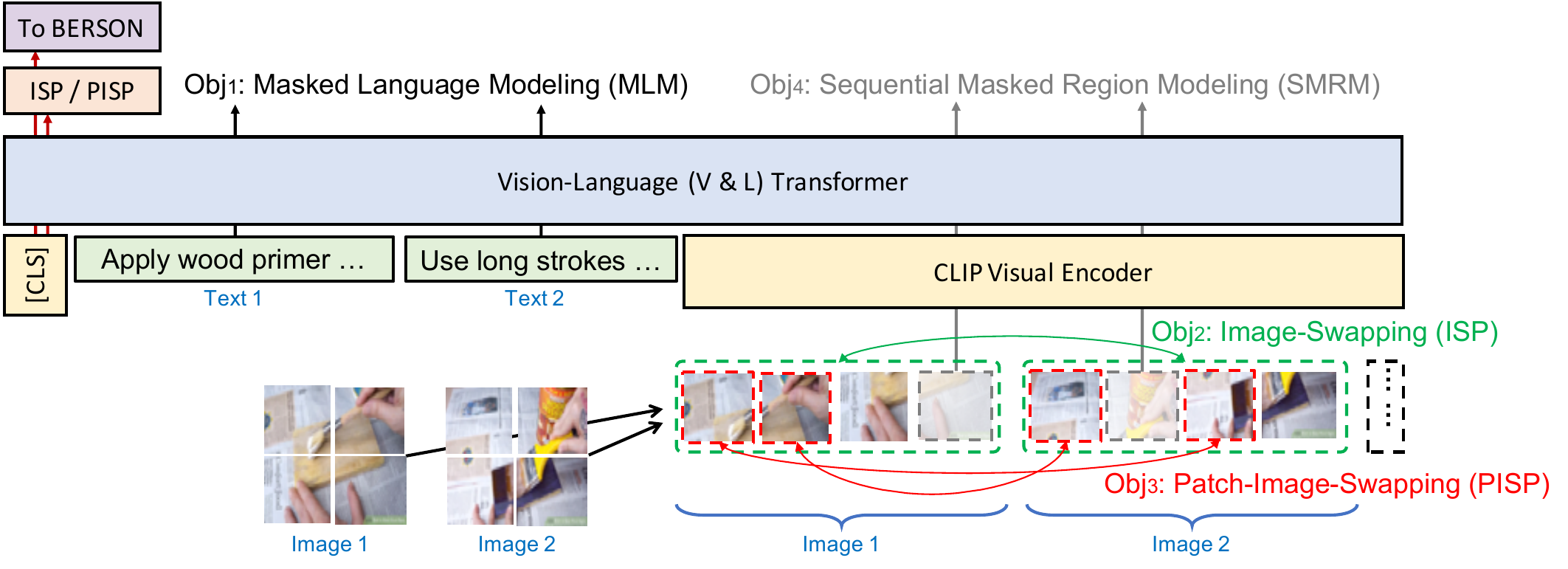}
    \vspace{-.5em}
    \caption{ \footnotesize
        \textbf{Sequence-aware pretraining} includes: (1) masked language modeling (MLM), (2) image-swapping prediction (ISP/PISP) which requires the model to predict if some images (image-patches) are swapped, and (3) sequential masked region modeling (SMRM) where models are asked to reconstruct masked regions in each image within the input sequence.
    }
    \label{fig:model_fig}
    % \vspace{-1em}
\end{figure*}

\mypar{Text-Only Encoders.}
We use \textit{RoBERTa}~\cite{liu2019roberta} for text-only inputs. Although the next-sentence prediction in BERT~\cite{devlin2019bert} can potentially be exploited for sequencing, we empirically find that RoBERTa % however not trained with such an objective,
performs better. % than BERT. , an optimized variant of BERT model~\cite{devlin2019bert},

\vspace{.3em}

\mypar{Multimodal Encoders.}
We consider the following two V\&L models mainly due to their easy adaptation to our proposed sequencing task: %their simplicity of adapting to our task:
% \footnote{\citet{bugliarello-etal-2020-multimodal} suggests that many V\&L models can achieve similar downstream performance if well trained.}

%\vspace{.3em}

\textit{VisualBERT}~\cite{li2019visualbert} grounds object detected image regions (\eg by Faster-RCNN~\cite{ren2016faster}) to language with a single transformer model~\cite{vaswani2017attention}. VisualBERT is pretrained with: (1) multimodal masked language modeling (MLM)\footnote{RoBERTa is used to initialize VisualBERT and CLIP-ViL.}, and (2) image-text matching prediction (ITM), where the image in an image-caption pair is randomly replaced with another one to create misalignment, and the model is required to predict whether the current pair is aligned.
% Typically, V\&L models use image captioning~\cite{chen2015microsoft} as training resource.

%\vspace{.3em}

\textit{CLIP-ViL}~\cite{shen2021much} is also a single-stream V\&L model similar to VisualBERT, while the visual encoder is replaced by a patch-based model inspired by the ViT~\cite{dosovitskiy2020image} in CLIP~\cite{radford2021learning}, where the image features are taken as \textit{gridded-image-patches} as shown in~\figref{fig:model_fig}.
The pretraining objectives remain the same as VisualBERT.
Empirically, both~\citet{shen2021much} and this work find such patch-based model tends to yield better downstream performance.
% and hence we mainly base our multimodal encoder off CLIP-ViL.

\vspace{.3em}

\mypar{Image-Only Encoders.} We attempt to provide an image-only baseline on our sequencing task with two visual encoders: (1) \textit{ResNet}-based~\cite{he2016deep} Faster-RCNN model (also the visual encoder in VisualBERT) where both the detected regional features and the whole-image-feature are used, and (2) the aforementioned patch-based \textit{CLIP} model.\footnote{Without confusion, throughout the paper we term the ViT- and CLIP-inspired visual encoder simply as CLIP.}
% Our empirical studies find that CLIP again shows better performance on the image-only version of our task, consistent with the multimodal version.

\subsection{Sequence-Aware Pretraining}

The standard multimodal grounding techniques~\cite{li2019visualbert, lu2019vilbert, su2019vl, chen2020uniter} do not explicitly concern the sequentiality of text and associated image sequences, and hence may fall short of effectively utilizing the sequential properties in multimodal inputs. 
To encourage models to have better awareness of the sequential alignments in multimodal instruction steps, we propose to pretrain the encoders with the following self-supervised objectives:
(1) masked language modeling (\textbf{MLM}),
(2) (patch-based) image-swapping predictions (\textbf{ISP/PISP}),
and (3) sequential masked region modeling (\textbf{SMRM}).
\figref{fig:model_fig} illustrates an overview of the pretraining paradigm.
% The proposed objectives can be applied to arbitrary length of an input sequence, we illustrate the proposed objectives with length $=2$ as it is sufficiently representative.

For the proposed objectives, the inputs to the models are generally \textit{ordered} instruction step sequences, which can be further sub-sampled to produce length-varying subsequences.
Although we do not find this necessarily benefit the downstream performance, it is observed that the sub-sampling helps the model converge faster.
While all of our proposed objectives can be applied to sequence with arbitrary length ($\geq2$),
without loss of generality and for simplicity, the following sections assume the sub-sampled sequence is of length $2$.

%\violet{I changed your obj name, please update the figure accordingly. \telin{done}}
\subsubsection{Masked Language Modeling}

The standard MLM~\cite{devlin2019bert} is employed by the text-only models to adapt a pretrained language model to the target domain (task instructions).
% which may contain articles outside of the pretraining domains, such as crafting manuals or recipes.
Following prior V\&L works, we apply MLM to multimodal models.
Specifically, we ensure that the textual description of each step $T_i$ gets similar amount of tokens being masked-out such that the models can potentially exploit the image sequences more.\footnote{As higher chances that the complementary textual information is also masked out from different steps.}

\subsubsection{Swapping-Based Prediction}
\label{sec:isp}

This objective concerns, with certain probability, randomly swapping a pair of items in a sequence and asking the model to judge whether the resulting sequence is properly ordered or not (\ie binary classification). We mainly perform the swapping in the image modality and hence it can be viewed as a sequence-aware version of ITM objective in most V\&L models.
As in ITM, the output representation at the \texttt{[CLS]} token is used to make the prediction. %\violet{what's ITM?}
% (with an MLP added on top).

\vspace{.3em}

\mypar{Standard.} 
For an ordered sequence $S$, we can randomly swap two\footnote{Two is our minimum number for a valid subsequence.} items of $S$, $\{S_i, S_j\}$, where $i < j$, to $\{S_j, S_i\}$, with a certain probability $\delta$.
Our preliminary studies find that swapping the textual contents does not necessarily help the downstream performance for either text-only or multimodal models, so we only perform the swapping on the images $\{I_i, I_j\}$ in both multimodal and image-only models.
For patch-based image inputs (or regional features), the whole patches of an image are swapped with those of another one within the same sequence, as illustrated in \textbf{Obj$_2$} in~\figref{fig:model_fig}.

\vspace{.3em}

\mypar{Patch-Based.}
We can perform the aforementioned swapping prediction with a finer granularity, directly on the image patches.
Assuming each image $I_i$ is cropped into $w$ patches (or $w$ detected regions), \ie $\{\textbf{i}_{i,k}\}_{k=1}^{w} = \{\textbf{i}_{i,1}, ..., \textbf{i}_{i,w}\}$, we randomly select $M$ (ranging from 1 to $w$) number of patches each from the two images $I_i, I_j$ (\ie $\{\textbf{i}_{i,p}\}, \{\textbf{i}_{i,q}\}, p, q \in M\text{-sized sampled indices}$) to be swapped with probability $\delta$. Specifically, for each image patch $\textbf{i}_{i,m} \in I_i$, a randomly selected image patch $\textbf{i}_{j,n} \in I_j$ is sampled to be swapped with.
% Note that we do not repeatedly sample the same $n$ for different $m$, and hence the swapping pair between $I_i, I_j$ is bijective.
The sampled $M$-sized indices do not need to be the same set of integers for each image. The \textbf{Obj$_3$} in~\figref{fig:model_fig} illustrates the patch-based swapping prediction with $w=4$ and $M=2$.

% \vspace{.3em}

\subsubsection{Sequential Masked Region Modeling}
Prior works extend the masked learning to the visual modality, where the masked target is either a predefined discrete visual vocabulary~\cite{sun2019videobert, bao2021beit} or (soft) object class labels~\cite{lu2019vilbert, su2019vl, chen2020uniter}. In this work, we construct a feature-based target vocabulary dynamically in each training batch.
We first randomly select the same amount of $X$\% ($X=15$) patches for \textbf{each image} to be masked out (replaced with 0-tensor), and then construct a target vocabulary from the original output representations (before masking) of these patches.

Concretely, denote the output representation of an input image-patch $\textbf{i}_{i,m}$
% through the visual encoder (ROI-Pooler or CLIP) and the subsequent multimodal transformer
as $h(\textbf{i})_{i,m}$ and the masked positions of $I_i$ as $D_i$, we can construct a candidate list from all the output representations of the patches at the masked positions of each image, \ie $C = \{h(\textbf{i})_{i,m}\} \cup \{h(\textbf{i})_{j,n}\}, m, n \in D_i, D_j$.
Denote the masked image patches (the gray-colored image patches in~\figref{fig:model_fig}) as $\textbf{mask(i)}_{i,m}$, for each output masked representation $h(\textbf{mask(i)})_{i,m}$, we concatenate it with all the candidates, \ie $h(\textbf{mask(i)})_{i,m} || h(\textbf{i'}), \forall \textbf{i'} \in C$, which results in $|C|$ concatenated representations for each masked position.
A $|C|$-way multi-class classification can then be performed by maximizing the probability of $p(\textbf{i}_{i,m} | h(\textbf{mask(i)})_{i,m}; C)$.
For robust training, we additionally:
(1) shuffle the candidate set $C$ for each masked position to prevent overfitting, 
and (2) ensure the overlapping of masked positions in each pair of images, $D_i \cap D_j$, is~\textless~50\%, allowing the models to utilize information of similar regions from other images in the sequence.
% to enhance the sequential awareness.

% \subsubsection{Others..., TCN, Margin loss, etc.}

\subsubsection{Overall Training Objective}
As the mechanism in some objectives cannot guarantee mutually exclusive impacts (\eg performing ISP and PISP simultaneously may create confusing swapped patches), we employ a turn-taking fashion, with uniform probability, one of the objectives (\textbf{Obj}) is sampled for each training mini-batch.
The overall pretraining objective is defined as below:
\begin{equation}
L = L_{\text{MLM}} + L_{\textbf{Obj}}, \footnotesize{\textbf{Obj} \sim \{\text{ISP}, \text{PISP}, \text{SMRM}\}}
\end{equation}

\subsection{Order Decoder -- BERSON}

% \subsubsection{Topological Sort}
% \telin{Maybe can omit if no longer important.}

% \subsubsection{BERSON}
BERSON is a recently proposed state-of-the-art neural sentence ordering framework~\cite{cuietal2020bert}, where a pointer network~\cite{vinyals2015order} exploits both the local (relative pairwise order) and global (self-attentions on top of the entire input sequence) information of the inputs to decode the predicted order.
BERSON mainly exploits the \texttt{[CLS]} output representations for relational understanding, which aligns well with how our encoders are pretrained (\figref{fig:model_fig}).
We integrate our encoders (with or without sequence-aware pretraining) into BERSON, replacing its original BERT encoder.
The BERSON-module-specific components are freshly initialized and then the entire integrated module is finetuned on our sequencing task.

\section{Experiments and Analysis}

Our experiments seek to answer these questions:
(1) How valid is the proposed task for humans to complete?
(2) Is multimodality helpful?
(3) Can the proposed sequence-aware pretraining utilize multimodality more effectively?
(4) How would results differ when alternative orders are considered?

\definecolor{LightCyan}{rgb}{0.88,1,1}
\newcolumntype{a}{>{\columncolor{LightCyan}}c}

\begin{table*}[th!]
\centering
\small
\scalebox{.82}{
    \begin{tabular}{@{ }c@{\ \ }rc|cccccc|cccccc@{ }}
    \toprule
    \multicolumn{1}{c}{\multirow{2}{*}{\textbf{Modality}}} & \multicolumn{1}{c}{\multirow{2}{*}{\textbf{Encoders}}} &
    \multicolumn{1}{c}{\textbf{Sequence-aware}} 
    & \multicolumn{6}{c}{\textbf{WikiHow Golden-Test-Set}} & \multicolumn{6}{c}{\textbf{RecipeQA Golden-Test-Set}} \\
    & & \textbf{Pretraining} & Acc$\uparrow$ & PMR$\uparrow$  & L$_{q}\uparrow$ & L$_{r}\uparrow$ & $\tau\uparrow$ & Dist$\downarrow$ & Acc$\uparrow$ & PMR$\uparrow$ & L$_{q}\uparrow$ & L$_{r}\uparrow$ & $\tau\uparrow$ & Dist$\downarrow$ \\
    \midrule
    
    % Human Test Set.
    
    \multirow{4}{*}{Image-Only}
    & \multirow{1}{*}{ResNet} & N
      & 21.73  & 2.00 & 2.81 & 1.73 & 0.01 & 7.87 & 31.20 & 5.00 & 3.27 & 2.07 & 0.27 & 6.10 \\
    \cline{2-15} \\[-.8em]
    & \multirow{1}{*}{CLIP} & N
      & 24.92 & 3.33 & 2.95 & 1.84 & 0.08 & 7.32 & 38.40 & 8.00  & 3.39 & 2.02 & 0.35 & 5.44 \\
    & \multirow{1}{*}{CLIP} & Y
      & 28.24 & 5.00 & 3.09 & 1.96 & 0.16 & 6.80 & 47.20 & 16.00 & 3.68 & 2.40 & 0.52 & 4.12 \\
    \cline{2-15} \\[-.8em]
    % {\color{blue}
    & \multicolumn{2}{a|}{\multirow{1}{*}{Human Performance}}
      & \multicolumn{1}{a}{68.16} & \multicolumn{1}{a}{47.49} & \multicolumn{1}{a}{4.27} & \multicolumn{1}{a}{3.51} & \multicolumn{1}{a}{0.72} & \multicolumn{1}{a}{2.43} & \multicolumn{1}{|a}{80.40} & \multicolumn{1}{a}{64.50} & \multicolumn{1}{a}{4.54} & \multicolumn{1}{a}{4.02} & \multicolumn{1}{a}{0.86} & \multicolumn{1}{a}{1.29} \\
    % }
    \midrule
    
    \multirow{3}{*}{Text-Only}
    & \multirow{1}{*}{RoBERTa} & N
      & 74.75 & 56.67 & 4.47 & 3.78 & 0.82 & 1.71 & 74.00 & 52.00 & 4.45 & 3.68 & 0.83 & 1.64 \\
    & \multirow{1}{*}{RoBERTa} & Y
      & 75.68 & 58.67 & 4.50 & 3.87 & 0.82 & 1.69 & 77.00 & 57.00 & 4.49 & 3.81 & 0.84 & 1.48 \\
    \cline{2-15} \\[-.8em]
    & \multicolumn{2}{a|}{\multirow{1}{*}{Human Performance}}
      & \multicolumn{1}{a}{83.35} & \multicolumn{1}{a}{66.91} & \multicolumn{1}{a}{4.63} & \multicolumn{1}{a}{4.11} & \multicolumn{1}{a}{0.89} & \multicolumn{1}{a}{1.06} & \multicolumn{1}{|a}{88.92} & \multicolumn{1}{a}{78.56} & \multicolumn{1}{a}{4.76} & \multicolumn{1}{a}{4.41} & \multicolumn{1}{a}{0.93} & \multicolumn{1}{a}{0.70} \\
    \midrule

    \multirow{5}{*}{Multimodal}
    & \multirow{1}{*}{VisualBERT} & N
      & 75.30 & 57.33 & 4.45 & 3.83 & 0.81 & 1.65 & 76.20 & 58.00 & 4.49 & 3.85 & 0.83 & 1.58 \\
    & \multirow{1}{*}{VisualBERT} & Y
      & 77.30 & 59.67 & 4.50 & 3.86 & 0.83 & 1.58 & 78.20 & 60.00 & 4.56 & 3.91 & 0.85 & 1.44 \\
    \cline{2-15} \\[-.8em]
    & \multirow{1}{*}{CLIP-ViL} & N
      & 76.15 & 59.00 & 4.49 & 3.87 & 0.82 & 1.68 & 79.20 & 60.00 & 4.57 & 3.93 & 0.85 & 1.29 \\
    & \multirow{1}{*}{CLIP-ViL} & Y
      & \textbf{79.87} & \textbf{65.67} & \textbf{4.57} & \textbf{4.05} & \textbf{0.85} & \textbf{1.44} & \textbf{82.60} & \textbf{68.00} & \textbf{4.61} & \textbf{4.10} & \textbf{0.88} & \textbf{1.10} \\
    \cline{2-15} \\[-.8em]
    & \multicolumn{2}{a|}{\multirow{1}{*}{Human Performance}}
      & \multicolumn{1}{a}{91.03} & \multicolumn{1}{a}{79.61} & \multicolumn{1}{a}{4.78} & \multicolumn{1}{a}{4.46} & \multicolumn{1}{a}{0.94} & \multicolumn{1}{a}{0.52} & \multicolumn{1}{|a}{92.12} & \multicolumn{1}{a}{83.13} & \multicolumn{1}{a}{4.82} & \multicolumn{1}{a}{4.53} & \multicolumn{1}{a}{0.95} & \multicolumn{1}{a}{0.45} \\

    \bottomrule
    
    \end{tabular}
}
\vspace{-.5em}
\caption{
\footnotesize
\textbf{Golden-test-set performance:} Models which take multimodal inputs (for both VisualBERT and CLIP-ViL encoders) consistently outperform the ones that only take unimodal inputs. Our proposed sequence-aware pretraining is shown consistently helpful throughout the three modality variants.
Humans show larger performance gain when both modalities of inputs are provided, and are more robust to the local ordering as implied by the smaller gaps between L$_{q}$ and L$_{r}$.
}
\label{tab:model_performances}
% \vspace{-.5em}
\end{table*}

\begin{table*}[th!]
\centering
\small
\scalebox{.85}{
    \begin{tabular}{cc|cccccc|cccccc}
    \toprule
    \multicolumn{1}{c}{\multirow{2}{*}{\textbf{Modality}}} &
    \multicolumn{1}{c}{\multirow{2}{*}{\textbf{Pretrain}}} 
    & \multicolumn{6}{c}{\textbf{WikiHow Golden-Test-Set}} & \multicolumn{6}{c}{\textbf{RecipeQA Golden-Test-Set}} \\
    & & Acc$\uparrow$ & PMR$\uparrow$  & L$_{q}\uparrow$ & L$_{r}\uparrow$ & $\tau\uparrow$ & Dist$\downarrow$ & Acc$\uparrow$ & PMR$\uparrow$ & L$_{q}\uparrow$ & L$_{r}\uparrow$ & $\tau\uparrow$ & Dist$\downarrow$ \\
    \midrule
    
    % Human Test Set.
    
    \multirow{2}{*}{Image-Only}
    & ISP
      & 27.31 & 4.00 & 3.02 & 1.82 & 0.12 & 7.00 & 43.20 & 9.00  & 3.49 & 2.05 & 0.47 & 4.46 \\
    & ISP + PISP
      & 27.57 & 4.67 & 3.07 & 1.93 & 0.16 & 6.85 & 43.40 & 12.00 & 3.57 & 2.24 & 0.48 & 4.46 \\
    % & MRM + IS + P-IS
    %   & 00.00 & 00.00 & 0.00 & 0.00 & 0.00 & 0.00 & 47.20 & 16.00 & 3.68 & 2.40 & 0.52 & 4.12 \\
    \midrule

    \multirow{5}{*}{Multimodal}
    & MLM
      & 77.08 & 61.33 & 4.52 & 3.96 & 0.83 & 1.65 & 79.60 & 61.00 & 4.55 & 3.93 & 0.86 & 1.29 \\
    & MLM + ISP
      & 77.61 & 62.00 & 4.54 & 3.97 & 0.83 & 1.60 & 80.00 & 61.00 & 4.56 & 3.93 & 0.86 & 1.26 \\
    & MLM + SMRM
      & 77.94 & 62.33 & 4.54 & 3.98 & 0.84 & 1.60 & 80.00 & 59.00 & 4.53 & 3.89 & 0.87 & 1.26 \\
    & MLM + ISP + PISP
      & 78.14 & 63.33 & 4.55 & 4.03 & 0.84 & 1.56 & 80.80 & 63.00 & 4.57 & 3.99 & 0.87 & 1.24 \\
    & MLM + ISP + SMRM
      & 79.47 & 63.67 & 4.57 & 4.03 & 0.85 & 1.54 & 81.40 & 63.00 & 4.57 & 4.00 & 0.87 & 1.20 \\
    % & MLM + IS + P-IS + MRM
    %   & 00.00 & 00.00 & 0.00 & 0.00 & 0.00 & 0.00 & 00.00 & 00.00 & 0.00 & 0.00 & 0.00 & 0.00 \\
    \bottomrule
    
    \end{tabular}
}
\vspace{-.5em}
\caption{
\footnotesize
\textbf{Model ablation studies:} We provide a performance breakdown for incremental combinations of the pretraining objectives, ablated on the best performing models (CLIP and CLIP-ViL) from~\tbref{tab:model_performances} for each dataset and modality.
}
\label{tab:ablation_studies}
% \vspace{-1.5em}
\end{table*}

\subsection{Evaluation Metrics}
% \telins{Needs some more thoughts on explaining why these metrics matter.}
% We adopt the following metrics from existing works for sentence ordering tasks:
% We use metrics following sentence ordering works:
We adopt metrics from sentence ordering works:
\vspace{.3em}
% \vspace{-.5em}

% \begin{itemize}[leftmargin=*]
% \item \textbf
\mypar{Position-Based} metrics concern the correctness of the absolute position of each item in a sequence, including:
(1) \textbf{Accuracy (Acc)} which computes the ratio of absolute positions in the ground truth order that are correctly predicted;
(2) \textbf{Perfect Match Ratio (PMR)} which measures the percentage of predicted orders exactly matching the ground truth orders;
and (3) \textbf{Distance (Dist.)} which measures the average distance\footnote{Except for distance metric, higher scores are better.} between the predicted and ground truth positions for each item.

% \vspace{-.5em}
\vspace{.3em}

% \item \textbf
\mypar{Longest Common Subsequence} computes the average longest subsequences in common~\cite{gong2016end} between the predicted and ground truth orders (\textbf{L$_{q}$}). We also consider a stricter version, longest common substring, which requires the consecutiveness for the comparisons (\textbf{L$_{r}$}).

% \vspace{-.5em}
\vspace{.3em}

% \item \textbf
\mypar{Kendall's Tau ($\tau$)}~\cite{lapata2003probabilistic} is defined as
$1 - 2\times (\#\ inversions) / (\#\ pairs)$, where the inversion denotes that the predicted relative order of a pair of items is inverted compared to the corresponding ground truth relative order, and $\#\ pairs = {N \choose 2}$ for $N$-length sequence.
% \end{itemize}

% \vspace{-.5em}
\vspace{.3em}

\noindent Each metric focuses on different perspectives of the predictions, \ie position metrics concern the absolute correctness, while common subsequence and $\tau$ metrics measure if general sequential tendency is preserved despite incorrect absolute positions.

\subsection{Implementation Details}

%\mypar{Training Details.}
We use the original data splits for RecipeQA.
For WikiHow, to prevent models' exploiting knowledge from similar articles, we split the data so that certain (sub)categories do not overlap in each split.
We use only the train splits in each dataset to perform their respective pretraining. 
More details of the data splits are in~\appendixref{a-sec:dataset_stats}.
% We select the model checkpoints to be evaluated using a held-out development split, specifically for WikiHow it is constructed from the same category distribution to the golden-test-set.
Preliminary studies show that joint training with both RecipeQA and WikiHow data does not necessarily improve the downstream performance, thus the models evaluated in the two datasets are trained simply using their respective training sets for faster convergence.

%\vspace{.3em}

%\mypar{Implementation Details.}
We cap the overall sequence length at $5$ and each step description with maximally $5$ sentences for both models and humans.
The maximum input length per step is $60$ tokens (overall maximum length $=300$) for training and GPU memory efficiency.
$\delta = 0.5$ for both ISP and PISP.
All images are resized to $224\times224$, and $32\times32$ patch is used for CLIP-based models, resulting in $7\times7=49$ patches per image.
Aside from standard positional embedding, we only supplement a modality token type embedding (text$:=$0, image$:=$1) to the multimodal models.
Pretrained weights for each encoder is obtained either from their corresponding code bases or by running their codes on our setup.\footnote{We initialize CLIP-ViL with our pretrained CLIP.}

% \vspace{-.5em}
\subsection{Standard Benchmark Results}
\label{sec:benchmark_results}
% \vspace{-.3em}

% \mypar{Human Performance.} The human performance for each input modality is reported in~\tbref{tab:model_performances}, where multimodal information is shown helpful consistently across the two datasets.

\noindent \tbref{tab:model_performances} summarizes both the human and model performance for each input modality evaluated using the original ground truth orders on the golden-test-set, whereas \tbref{tab:ablation_studies} summarizes a more detailed breakdown of the model performance when incrementing combinations of pretraining objectives.

As is shown, multimodal information is verified consistently helpful for humans.
Compared under same scenario with or without the sequence-aware pretraining, the two multimodal models consistently outperform their text-only counterparts, where the proposed pretraining technique is shown particularly effective for the patch-based multimodal model (CLIP-ViL).
However, our top-performing models still exhibit significant gaps below human performance, especially in PMR.

Additionally, we observe a different trend in the two datasets where the multimodality benefits more in RecipeQA than WikiHow.
The gap between the multimodal human and model performance is larger than the text-only counterparts in WikiHow, while a reversed trend is shown in RecipeQA.
We hypothesize that recipes may contain more domain-specific language usages and/or less words for the pretrained language models and hence benefits more from the our in-domain sequence-aware pretraining.
Humans, on the other hand, benefit more from the images in WikiHow as its texts are hypothesized to contain more ambiguities.

\vspace{.3em}

\begin{figure}[t!]
\centering
    \includegraphics[width=1.0\columnwidth]{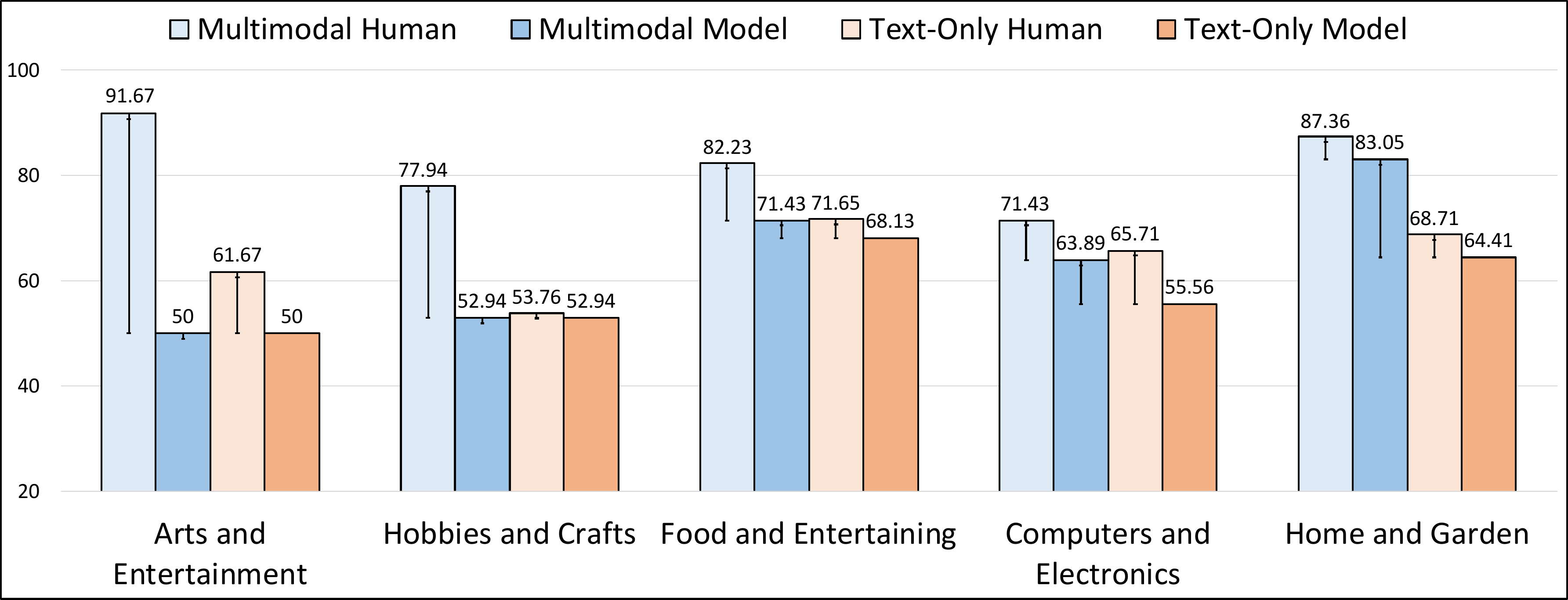}
    \caption{
    \footnotesize
    \textbf{Top-3 and least-2 categories of \textbf{human-model} performance difference (in PMR):} The selected categories have~\textgreater10 samples. The difference bars on the multimodal model series are compared against the text-only model series.
    }
    \label{fig:model_human_cats}
    % \vspace{-1.5em}
\end{figure}

\mypar{WikiHow Category Analysis.}
% \mypar{Ablation Studies.}
% \tbref{tab:ablation_studies} summarizes a more detailed breakdown of the performance when different combinations of pretraining objectives are applied.
% \telin{Write more...}
% \vspace{.3em}
% \input{figs/top_3_categories_uni_multi}
% \mypar{Category-Wise Performance.}
We are interested in knowing on which categories of WikiHow our models perform closer to humans, and on which the multimodal information is most efficiently utilized.
In~\figref{fig:model_human_cats} we select categories with the top and least performance gaps (with PMR metric, top=3, least=2) between the human and our best performing models.
% , and in~\figref{fig:uni_multi_cats} we show performance gaps between the best performing multimodal and text-only models.
We observe that the categories on which multimodal models outperform the text-only ones the most are also the categories the models perform closest to humans, \eg \textit{Home and Garden}.
We hypothesize that the images in these categories are well complementary to the texts and that our sequence-aware grounding performs effectively.
In contrast, in categories such as \textit{Arts and Entertainment} and \textit{Hobbies and Crafts} where humans still enjoy benefits from multimodal information, our models have difficulty utilizing the multimodal information.
We hypothesize that better visual understanding may alleviate the potentially suboptimal grounding as images of these categories can contain many non-common objects.
% (\ie more social and personal).
% The observation aligns with our intuitions that models benefit from multimodality on tasks that follow a definite order (\ie procedural data, like a physical category), compared with tasks that do not, where visual information may, on the contrary, be misleading.
% For examples, see~\appendixref{a-sec:quali}.

\begin{table*}[th!]
\centering
\small
\scalebox{.89}{
    \begin{tabular}{cr|cccccc|cccccc}
    \toprule
    \multicolumn{1}{c}{\multirow{3}{*}{\textbf{Modality}}}
    & \multicolumn{1}{c|}{\multirow{3}{*}{\textbf{Subset}}}
    & \multicolumn{6}{c|}{\textbf{WikiHow Golden-Test-Set (Size: 300)}}
    & \multicolumn{6}{c}{\textbf{RecipeQA Golden-Test-Set (Size: 100)}} \\
    &
    & \multicolumn{2}{c}{Acc$\uparrow$} & \multicolumn{2}{c}{PMR$\uparrow$} & \multicolumn{2}{c|}{L$_{r}\uparrow$}
    & \multicolumn{2}{c}{Acc$\uparrow$} & \multicolumn{2}{c}{PMR$\uparrow$} & \multicolumn{2}{c}{L$_{r}\uparrow$} \\
    &
    & single & multi & single & multi & single & multi
    & single & multi & single & multi & single & multi \\
    
    \midrule

    \multirow{8}{*}{Text-Only}
    & \multirow{1}{*}{Single}
      & 77.30 & --- & 61.75 & --- & 3.98 & --- & 79.32 & --- & 60.23 & --- & 3.90 & --- \\
    & \multirow{2}{*}{Multi.}
      & 67.35 & 80.00 & 40.82 & 59.18 & 3.35 & 3.86 & 60.00 & 75.00 & 33.33 & 58.33 & 3.17 & 3.92 \\
    & & \multicolumn{6}{c|}{(\% of instances benefit w. multi-reference: 34.7\%)} & \multicolumn{6}{c}{(\% of instances benefit w. multi-reference: 50.0\%)} \\
    & \multirow{1}{*}{All}
      & 75.68 & 77.74 & 58.67 & 61.67 & 3.87 & 3.96 & 77.00 & 78.80 & 57.00 & 60.00 & 3.81 & 3.90 \\
    \cline{2-14} \\[-.8em]
    & \multirow{1}{*}{Single$\dagger$}
      & 85.57 & --- & 71.41 & --- & 4.24 & --- & 90.27 & --- & 80.41 & --- & 4.47 & --- \\
    & \multirow{2}{*}{Multi.$\dagger$}
      & 72.03 & 85.51 & 43.84 & 71.38 & 3.46 & 4.14 & 79.00 & 87.00 & 65.00 & 80.00 & 3.95 & 4.40 \\
    & & \multicolumn{6}{c|}{(\% of instances benefit w. multi-reference: 42.9\%)} & \multicolumn{6}{c}{(\% of instances benefit w. multi-reference: 41.6\%)} \\
    & \multirow{1}{*}{All$\dagger$}
      & 83.35 & 85.56 & 66.91 & 71.40 & 4.11 & 4.22 & 88.92 & 89.88 & 78.56 & 80.36 & 4.41 & 4.46 \\

    \midrule

    \multirow{8}{*}{Multimodal}
    & \multirow{1}{*}{Single}
      & 81.68 & --- & 69.90 & --- & 4.15 & --- & 83.71 & --- & 69.07 & --- & 4.12 & --- \\
    & \multirow{2}{*}{Multi.}
      & 70.98 & 78.82 & 47.05 & 61.22 & 3.59 & 3.90 & 46.67 & 60.00 & 33.33 & 33.33 & 3.67 & 3.78 \\
    & & \multicolumn{6}{c|}{(\% of instances benefit w. multi-reference: 21.6\%)} & \multicolumn{6}{c}{(\% of instances benefit w. multi-reference: 66.6\%)} \\
    & \multirow{1}{*}{All}
      & 79.87 & 81.19 & 65.67 & 68.00 & 4.05 & 4.11 & 82.60 & 83.00 & 68.00 & 68.00 & 4.10 & 4.11 \\
    \cline{2-14} \\[-.8em]
    & \multirow{1}{*}{Single$\dagger$}
      & 92.86 & --- & 83.67 & --- & 4.56 & --- & 91.88 & --- & 82.61 & --- & 4.52 & --- \\
    & \multirow{2}{*}{Multi.$\dagger$}
      & 82.09 & 92.22 & 59.80 & 83.33 & 3.99 & 4.54 & 100.00 & 100.00 & 100.00 & 100.00 & 5.00 & 5.00 \\
    & & \multicolumn{6}{c|}{(\% of instances benefit w. multi-reference: 41.18\%)} & \multicolumn{6}{c}{(\% of instances benefit w. multi-reference: 0.0\%)} \\
    & \multirow{1}{*}{All$\dagger$}
      & 91.03 & 92.75 & 79.61 & 83.61 & 4.46 & 4.55 & 92.12 & 92.12 & 83.13 & 83.13 & 4.53 & 4.53 \\

    \bottomrule
    \end{tabular}
}
\footnotesize{$\ast$ The size of the \textbf{Multi.} subsets in (\textit{text-only}, \textit{multimodal}) are: (49, 51)/300 in WikiHow and (12, 3)/100 in RecipeQA.}
\vspace{-.5em}
\caption{
\footnotesize
\textbf{Multi-reference performance:} ($\dagger$ denotes human performance)
Our golden-test-set can be decomposed into two subsets: \textbf{Single} where each instance in this subset only has one single originally authored ground truth, and \textbf{Multi.} where each instance features multiple ground truths from alternative orders.
For the \textbf{Multi.} subset, two types of performance can be computed: \textbf{single} considers only the originally authored ground truth and \textbf{multi} computes the multi-reference performance.
\textbf{All} denotes the entire test-set combining the results from \textbf{Single} and \textbf{Multi.} subsets.
% For each \textbf{subset} featuring a single ground truth reference (\textbf{Single}) or multiple ground truth references (\textbf{Multi.}), we can compute two types of performance: one considers only the original authored ground truth (\textbf{single}) and the other computes the multi-reference performance for those instances annotated with alternative orders (\textbf{multi}).
Results are reported on the two main competitors: multimodal and text-only using the best performing models from~\tbref{tab:model_performances} in each modality.
\textbf{\% of instances benefit w. multi-reference} indicates that of what percentage of instances \textit{in each multi-reference subset} humans and the models benefit (for each instance if its performance improves \textit{in any of the metrics}) from alternative ground truth orders.
}
\label{tab:multiref_perf}
\vspace{-1em}
\end{table*}

% \vspace{-.3em}
\subsection{Evaluating with Alternative Orders}
\label{sec:eval_alternative_orders}
% \vspace{-.3em}

For each instance where alternative ground truth orders exist, the performance is computed by the best each predicted order can obtain against all the ground truth orders\footnote{Jointly considered from all the evaluation metrics.}, denoted by \textbf{multi-reference performance}, and the subset containing these instances is denoted as the \textbf{multi-reference subset}.\footnote{The overall average number of ground truth references becomes 1.19, 1.23, 1.09 for multimodal, text-only, and image-only versions in WikiHow; and 1.10, 1.17, 1.14 in RecipeQA.}

\vspace{.3em}

\mypar{Statistics.}
\tbref{tab:alternative_stats} lists the essential statistics of the multi-reference subsets, including the counts of the multi-reference instance for each dataset and modality, as well as the per-instance statistics.

\vspace{.3em}

\mypar{Multi-Reference Performance.}
The noticeable main competitors in~\tbref{tab:model_performances} are multimodal and text-only models, and hence for conciseness, in~\tbref{tab:multiref_perf} we mainly report the multi-reference version of their best performing variants with the selected metrics.
Several trends still hold:
(1) Multimodal models still outperform the text-only counterparts.
(2) Human performance is still well above models' even under multi-reference setups.
Additionally, both humans and models perform significantly worse in the multi-reference subset when single (original) ground truth is enforced, implying the validity of our alternative order annotations. 

We originally hypothesize that enforcing the original authored order to be the only ground truth would be unfair to the text-only models, as images can often better represent the detailed scene changes omitted by the texts, while in reality certain steps may not need to strictly follow the authored order.
Judging from the number of instances that improve after evaluating with alternative orders, the text-only model indeed benefits more from the multi-reference setup.
Examining the general trends in~\tbref{tab:multiref_perf}, one can conclude that the textual contents indeed posses certain levels of ambiguities where images can help to alleviate. However, as the performance gaps between multimodal and text-only models are still significant under the multi-reference settings, %enforcing the original order as the only ground truth should not be the major reason justifying 
advantages of multimodality. 
Note that humans achieve perfect performance on the multi-reference subset in RecipeQA, though unlikely it may seem, it is mainly due to recipes tend to have rarer possible alternative orders.

\vspace{.3em}

\begin{table}[t!]
\centering
\footnotesize

\scalebox{0.79}{
    \begin{tabular}{c|ccc|ccc}
    \toprule
    \multicolumn{1}{c}{\multirow{2}{*}{\textbf{Modality}}}
    & \multicolumn{3}{c}{\textbf{WikiHow (300)}}
    & \multicolumn{3}{c}{\textbf{RecipeQA (100)}} \\
    & Cnt & Min/Max & Avg/Std & Cnt & Min/Max & Avg/Std \\
    \midrule % \\[-1.1em]
    
    \multirow{1}{*}{Image-Only}
    & 24 & 2/4 & 2.1/1.4 & 13 & 2/3 & 2.1/0.3 \\
    \midrule
    
    \multirow{1}{*}{Text-Only}
    & 49 & 2/6 & 2.4/0.9 & 12 & 2/6 & 2.4/1.1 \\
    \midrule
    
    \multirow{1}{*}{Multimodal}
    & 51 & 2/4 & 2.1/0.5 & 3 & 2/6 & 4/1.6 \\
    
    \bottomrule
    \end{tabular}
}

\iffalse
\scalebox{0.79}{
    \begin{tabular}{cr|cccccc}
    \toprule
    \multicolumn{1}{c}{\multirow{2}{*}{\textbf{Modality}}} & \multicolumn{1}{c}{\multirow{2}{*}{\textbf{Refs.}}} &
    \multicolumn{6}{c}{\textbf{WikiHow Multi-Ref. Subset}} \\
    & & Acc$\uparrow$ & PMR$\uparrow$  & L$_{q}\uparrow$ & L$_{r}\uparrow$ & $\tau\uparrow$ & Dist$\downarrow$ \\
    \midrule % \\[-1.1em]
    
    \multirow{2}{*}{Image-Only}
    & \multirow{1}{*}{Single}
      & 00.00 & 00.00 & 0.00 & 0.00 & 0.00 & 0.00 \\
    & \multirow{1}{*}{Multi.}
      & 00.00 & 00.00 & 0.00 & 0.00 & 0.00 & 0.00 \\
    % \cline{2-9} \\[-.8em]
    \midrule

    \multirow{2}{*}{Text-Only}
    & \multirow{1}{*}{Single}
      & 00.00 & 00.00 & 0.00 & 0.00 & 0.00 & 0.00 \\
    & \multirow{1}{*}{Multi.}
      & 00.00 & 00.00 & 0.00 & 0.00 & 0.00 & 0.00 \\
    % \cline{2-9} \\[-.8em]
    \midrule
    
    \multirow{2}{*}{Multimodal}
    & \multirow{1}{*}{Single}
      & 00.00 & 00.00 & 0.00 & 0.00 & 0.00 & 0.00 \\
    & \multirow{1}{*}{Multi.}
      & 00.00 & 00.00 & 0.00 & 0.00 & 0.00 & 0.00 \\
    % \cline{2-9} \\[-.8em]
    
    \bottomrule
    \end{tabular}
}
\fi

\caption{
\footnotesize
\textbf{Multi-reference subset statistics:} We report the count (cnt) of multi-reference instances in each dataset across the three modalities, and their basic statistics.
}
%\vspace{-.8em}
\label{tab:alternative_stats}
\end{table}

\vspace{.3em}

\mypar{WikiHow Categories.}
\tbref{tab:alternative_mean_gt_by_cats} lists the WikiHow categories with the most (top-5) annotated multi-reference ground truths.
Note that the categories with more annotated alternative ground truths are also among the worse performance from both humans and models (\figref{fig:model_human_cats}).
We provide sample qualitative inspections in~\appendixref{a-sec:quali}.
% and \figref{fig:uni_multi_cats}).

% \input{tables/dataset_stats}

% \input{tables/multi_ref_benefits}

% \input{tables/alternative_order_modality_quad}

\begin{table}[t!]
\centering
\small
\scalebox{.9}{
    \begin{tabular}{r|ccc}
    \toprule
    \multirow{2}{*}{\textbf{Categories}} & \multicolumn{3}{c}{\multirow{1}{*}{\textbf{Mean Per-Instance Refs. (Cnt)}}} \\
                                         & \textbf{Multimodal} & \textbf{Text} & \textbf{Image} \\
    \midrule % \\[-1.1em]
    
    % Human Test Set.
    \multirow{1}{*}{Home and Garden}
    & 2.00 (7) & 2.14 (7) & 2.00 (3) \\
    \multirow{1}{*}{Hobbies and Crafts}
    & 2.00 (5) & 2.73 (11) & 2.00 (2) \\
    \multirow{1}{*}{Food and Entertaining}
    & 2.20 (15) & 2.22 (14) & 2.17 (12) \\
    \multirow{1}{*}{Others}
    & 2.28 (7) & 2.67 (5) & 2.00 (4)\\
    \multirow{1}{*}{Personal Care and Style}
    & 2.33 (3) & 2.00 (1) & 2.00 (1) \\

    \bottomrule
    \end{tabular}
}
\caption{
\footnotesize
\textbf{Top-5 mean alternative orders by categories:} We list top-5 categories in WikiHow according to the number of average ground truth references in their multi-reference subset. We again only list the categories with total instance count~\textgreater10.
}
\label{tab:alternative_mean_gt_by_cats}
\vspace{-1em}
\end{table}

\iffalse
\mypar{Zero-Shot Manual Completion.}
One potential downstream application for task-order sequencing is to infer the missing steps in an \textit{incomplete} instructional manual.
For each task in the human test set, we randomly \textit{skip a step} and provide the rest of the steps to a trained model in \textit{ground truth order}.
The model is then asked to retrieve the missing step from a step pool composed by all the steps in this dataset.
Using a trained model, we take the `[CLS]' output representations of both the incomplete manual and each step in the pool to perform a \textit{zero-shot} nearest neighbor search\footnote{We feed the incomplete manual at once to the retrieval model and mean pool the step-wise {\tt `[CLS]'} representations.}.
We compute the mean reciprocal rank (MRR) and Top-1 accuracy for such step retrieval.
Additionally, we evaluate whether the model can \textit{identify} which step is missing by correctly ``putting" the retrieved step back to \textit{complete} a proper manual, via the learned sequencing capability\footnote{Sequencing is performed with appending the retrieved step to the end of the incomplete manual.}.
We compute a \textit{mean reciprocal step recognition} (MRSR) as:
$\frac{\mathds{1}[\text{correct prediction}]}{\text{rank of gt}}$.
The results shown in~\tbref{tab:zero_shot} suggests that multimodality is likewise helpful in this downstream task.
\fi

\section{Related Work}
% \vspace{-.5em}

\mypar{Sequence Ordering.}
Story sequencing test is a popular way of examining children’s abilities on sequential reasoning
% of events and comprehending procedures,
which is shown evident for procedural understanding~\cite{tomkins1952tomkins, baron1986mechanical, loucks2017children}.
In NLP, existing works attempt the sequencing task as sorting a series of unordered sentences~\cite{chen2016neural, cui2018deep, logeswaran2018sentence, oh2019topic, lee2020slm, calizzano2021ordering} from paper abstracts or short paragraphs.
While certain prior work also attempts to extend it to incorporate multimodality~\cite{agrawal2016sort}, the dataset used, Visual StoryTelling~\cite{huang2016visual}, features album images that were not intended to be procedural nor supply unstated details to complement the texts.
% \footnote{The images come from photo albums and the annotators \textit{create} stories based on their freely arranged image sequences.}
% \citet{lee2020slm} shows learning to \textit{unshuffle} sentences help several downstream NLP tasks.
In computer vision, existing work leverages shuffle frame prediction for learning video representations~\cite{lee2017unsupervised, xu2019self, wang2020order, li2020hero} as well as cycle consistency constraints for learning temporal dynamics~\cite{epstein2021learning}.
~\newcite{zellers2021merlot} also features a pairwise relative frame re-ordering objective to learn temporal common sense from scripted videos, however, as their downstream tasks mainly concern visual reasoning and ordering by frame-text-matching (also on Visual StoryTelling), the re-ordering objective is more focused on the visual modality.
Our work takes a different perspective to tackle a comprehensive multimodal sequencing task with a focus on the procedural task-solving knowledge and gauging the helpfulness of complementary information in different modalities.
% , and establish the human benchmarks.

\vspace{0.3em}

\mypar{Task/Procedure Understanding.}
Other works have utilized WikiHow for learning task knowledge.
In NLP, textual descriptions of WikiHow have been used for abstractive summarization~\cite{koupaee2018wikihow}, procedural understanding~\cite{zhou-etal-2019-learning-household,tandon-etal-2020-dataset}, and intent estimation~\cite{zhang-etal-2020-intent}.
Prior work~\cite{zhang-etal-2020-reasoning} considers WikiHow for learning event temporal ordering, but limited to only pairwise relations.
A concurrent work uses WikiHow to infer visual goals~\cite{yang2021visual}.
We hope our curation can help advancing the goal of comprehensive multimodal procedural understanding.

Another popular form of comprehending given procedures is through a \textit{multiple choice} machine comprehension task. Prior work has utilized text book figures~\cite{kembhavi2017you} as a \textit{holistic} "reading reference" for models to select the correct order of certain (textually described) events from \textit{given multiple choices}. Another work attempts the original visual ordering task of RecipeQA~\cite{liu2020multi} (also an multiple choice task). However, we argue that our task tackles a more complex task as the desired orders need to be directly derived and the event-wise complementary multimodal understanding is not an essential component in these existing works.

\vspace{0.3em}

\mypar{Multimodality.}
Beside models used in this work, there are several recent advanced multimodal grounding techniques~\cite{tan2019lxmert, li2019visualbert, lu2019vilbert, su2019vl, chen2019uniter, huang2020pixel,wen2021cookie}.
We utilize VisualBERT and CLIP-ViL for their simplicity to be adapted to our task and easier integration to our proposed pretraining techniques, however, our framework is able to incorporate any of the aforementioned multimodal models.

% \vspace{-.3em}
\section{Conclusions}
% \vspace{-.5em}

In this work we present studies of language and multimodal models on  procedure sequencing, leveraging popular online instructional manuals.
Our experiments show that both multimodality and our proposed sequence-aware pretraining are helpful for multimodal sequencing, however, the results also highlight significant gaps below human performance ($\sim$ 15\% on PMR).

We provide insights as well as resources, such as the multi-reference annotations of the sequencing task, to spur future relevant research.
We also anticipate that the alternative orders defined and annotated in our work can benefit more comprehensive task-procedure understanding. 
Future work such as predicting task steps which can be parallel or interchangeable, and understanding step dependencies can be explored.

\section*{Acknowledgments}

Many thanks to Liunian Harold Li for his original CLIP-ViL implementation; to I-Hung Hsu and Zi-Yi Dou for their helpful discussion and feedback; and to the anonymous reviewers for their constructive suggestions.
This material is based on research supported by the Machine Common Sense (MCS) program under Cooperative Agreement
N66001-19-2-4032 with the US Defense Advanced Research Projects Agency (DARPA) and a CISCO research contract. %The U.S. Government is authorized to reproduce and distribute reprints for Governmental purposes notwithstanding any copyright notation thereon. 
The views and conclusions contained herein are those of the authors and should not be interpreted as necessarily representing %the official policies or endorsements, either expressed or implied, of 
DARPA, CISCO, or the U.S. Government.

\clearpage

\section*{Ethics and Broader Impacts}

We hereby acknowledge that all of the co-authors of this work are aware of the provided \textit{ACM Code of Ethics} and honor the code of conduct.
This work is mainly about sequencing a given series of multimodal task procedures, represented by text descriptions along with their images.
The followings give the aspects of both our ethical considerations and our potential impacts to the community.

\vspace{.3em}

\mypar{Dataset.}
We collect the human performance on our sequencing task (both the standard human performance and the alternative order annotations) via Amazon Mechanical Turk (MTurk) and ensure that all the personal information of the workers involved (e.g., usernames, emails, urls, demographic information, etc.) is discarded in our dataset.
While the sequence orders either from the original author intended ones or those annotated by the workers for the standard performance may possess unintended biases against certain population group of people (\eg due to cultural differences or educational differences, some tasks may be performed differently from the original intended orders), we anticipate the additional multi-reference annotation can alleviate such an issue as well as provide a broader view to approach procedural understanding, \ie certain task-steps can be interchanged.

This research has been reviewed by the \textbf{IRB board} and granted the status of an \textbf{IRB exempt}.
The detailed annotation process (pay per amount of work, guidelines) is included in the appendix; and overall, we ensure our pay per task is above the the annotator's local minimum wage (approximately \$12 USD / Hour).
We primarily consider English speaking regions for our annotations as the task requires certain level of English proficiency.

\vspace{.3em}

\mypar{Techniques.} We benchmark the proposed sequencing task with the state-of-the-art large-scale pretrained language and multimodal models with our novel sequence-aware pretraining techniques.
As commonsense and task procedure understanding are of our main focus, we do not anticipate production of harmful outputs, especially towards vulnerable populations, after training models on our proposed task.
% \clearpage

\bibliography{anthology,custom}
\bibliographystyle{acl_natbib}

\clearpage

\appendix

\section{Details of Datasets}
\label{a-sec:dataset_stats}

\subsection{Image Contents}

For simplicity and computational concerns, in this work we only pair one image to each of its associated task-step textual descriptions.
However, in both WikiHow and RecipeQA, each task-step can have more than one associated images or visual contents represented by short clips or GIFs.
We simply select the first image, which is supposed to be the most representative, for those step featuring multiple images; and sample the frame in the middle of time interval for clips or GIFs.
Nevertheless, our framework does not assume any limitation on how many images per step to be processed.

\subsection{WikiHow Categories}

The category in WikiHow generally forms a hierarchical directed acyclic graph.
Each category can have its relevant subcategory, which usually spans finer-granularity of category types.
For example, a possible category traversal path is: \textit{Cars and Vehicles} \textrightarrow \textit{Public Transport} \textrightarrow \textit{Air Travel}, which can lead to the article \textit{How to Overcome the Fear of Flying}.
We attach these full category traversal paths as an additional feature to each of the article in our dataset, and we also will provide a complete list of the taxonomy composed by all the categories and subcategories in WikiHow.
We include the category-data counts in~\tbref{tab:all_cats} for a reference, where we only show the top-level category here. The more in-depth categories can be referred to in the full released version of the dataset.

\subsection{Train-Dev Splits}
\label{a-sec:splits}

For RecipeQA we use the original data splits which ensure no identical recipe appears in more than one set (each recipe has its unique recipe-id), as this dataset only has one category and the data quality is much more uniform than that of WikiHow, \ie most recipes fulfill our target dataset criteria.

For WikiHow, we split the data according to the third level category to prevent models from exploiting too similar task knowledge in the same category, where the level (three) is empirically decided.
% This will ensure that no articles belonging to the same third-level category should appear in more than one set.
Specifically, we ensure that the third-level categories where the articles in our golden-test-set belong to, do not appear in the train set.
% We also prioritize categories that are more likely to be procedural (concerning more physical knowledge) for the test sets, such as \textit{Cars and Vehicles}, \textit{Hobbies and Crafts}, and \textit{Home and Garden}.
We first split the WikiHow dataset into train, development, and test set following this strategy, and then construct our golden-test-set by sub-sampling a subset of this (larger) test set followed by manual inspections, to ensure its quality.
And then, we simply join the remaining test set samples to the development set.
Refer to \tbref{tab:data-det} in the main paper for detailed statistics.

\begin{table}[t!]
\centering
\small
% \scalebox{.88}{
    \begin{tabular}{r|c}
    \toprule
    \textbf{Categories} & \textbf{Counts} \\
    \midrule % \\[-1.1em]
    Arts and Entertainment & 4675 \\
    Cars and Other Vehicles & 2044 \\
    Computers and Electronics & 15023 \\
    Education and Communications & 7406 \\
    Family Life & 1747 \\
    Finance and Business & 6228 \\
    Food and Entertaining & 7670 \\
    Health & 8800 \\
    Hobbies and Crafts & 9217 \\
    Holidays and Traditions & 736 \\
    Home and Garden & 9460 \\
    Personal Care and Style & 6523 \\
    Pets and Animals & 5281 \\
    Philosophy and Religion & 828 \\
    Relationships & 2877 \\
    Sports and Fitness & 3271 \\
    Travel & 746 \\
    Work World & 1579 \\
    Youth & 2389 \\
    Others & 21 \\
    \bottomrule
    \end{tabular}
% }
\caption{
\footnotesize
\textbf{Top-Level Categories of WikiHow:} Number of unique articles in each top-level category of the WikiHow dataset. The categories are sorted by alphabetical order. In total there are 19 top-level categories (same as what this page indicates: https://www.wikihow.com/Special:CategoryListing), and one "others" category for standalone leaf nodes without real linkages to these top-level categories.
}
\label{tab:all_cats}
\end{table}

\section{Details of Human Annotation}
\label{a-sec:human}

\subsection{Golden-Test-Set Selections}

In order to construct a high-quality test set for humans to evaluate, we manually select the samples which meet our general criteria: (1) the tasks are procedural in both texts and images (2) the task's images are designed to complement the textual descriptions or provide a more illustrative information for some unstated implicit knowledge. We ask three of our internal members (co-authors) to perform such manual selection, and preserve ones that have majority votes.
In total, we select 300 samples for WikiHow and 100 samples for RecipeQA.

\subsection{General Annotation Procedure}

\subsubsection{Standard Performance Benchmark}

\begin{figure*}[t!]

\begin{subtable}{\textwidth}
\centering
\centering
    \includegraphics[width=.95\columnwidth]{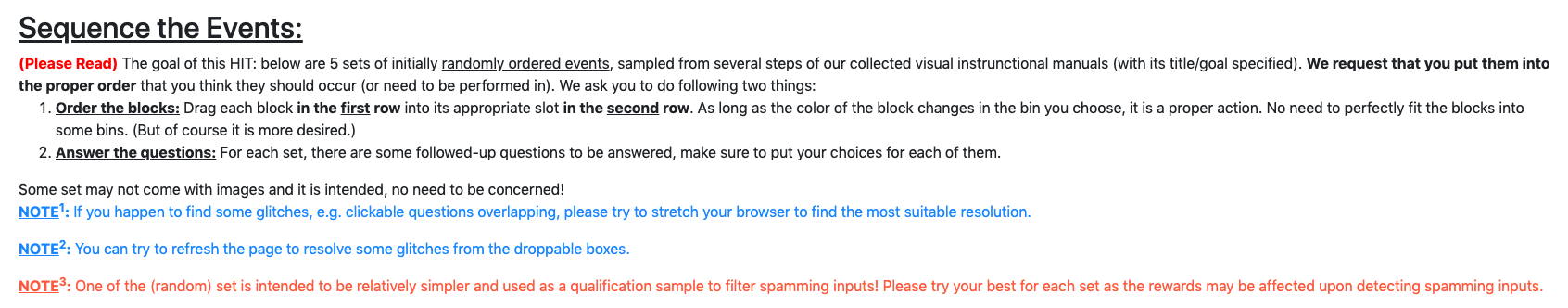}
\caption{\footnotesize Human Annotation Instruction}
\label{fig:ui_instr}
\end{subtable}

\begin{subtable}{\textwidth}
\centering
\centering
    \includegraphics[width=.95\columnwidth]{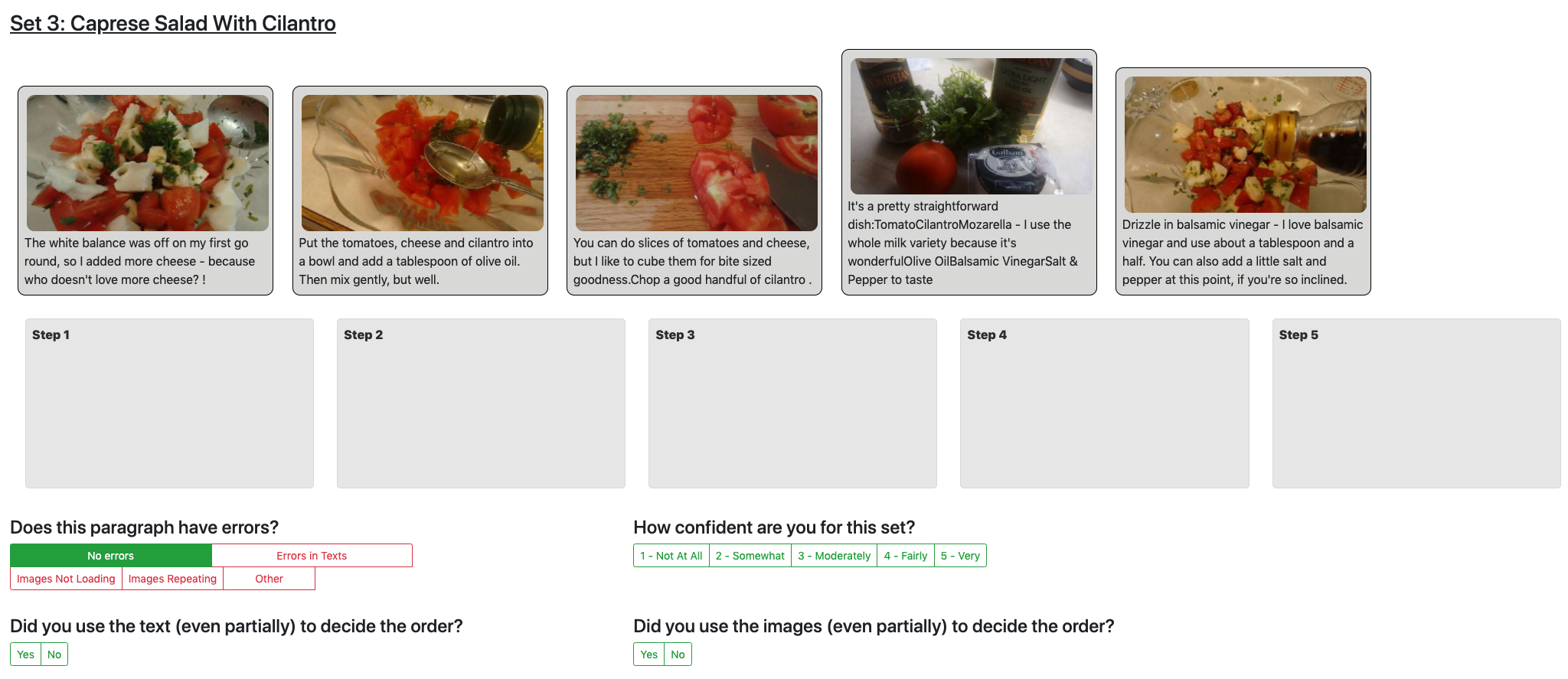}
\caption{\footnotesize Sample Annotation Interface}
\label{fig:ui_ui}
\end{subtable}

\vspace{.5em}

\caption{ \footnotesize
    \textbf{MTurk Annotation User Interface:}
    \textbf{(a)} We ask the annotator to follow the indicated instruction, and perform the sequencing task.
    \textbf{(b)} The annotation task is designed for an intuitive \textit{drag-and-drop} usage, followed by a few additional questions such as confidence level and whether each modality helps.
    (This example is obtained from RecipeQA dataset.)
}
\label{fig:ui}
\end{figure*}

We collect the human performance via Amazon Mechanical Turk (MTurk).
Each MTurk worker is required to read the provided instruction carefully, as shown in~\figref{fig:ui_instr}, and then perform the task, which is designed to be done in an intuitive \textit{drag-n-drop} (illustrated in~\figref{fig:ui_ui}) fashion.

Each MTurk HIT is designed to have five sets of sequencing tasks followed by a few additional questions such as confidence level of the worker when inferring the order, and whether different modalities are helpful in a particular task.
For each unique sample in the selected golden-test-set, we construct three annotation sets each for one modality version: multimodal, text-only, and image-only.
We launch the HITs containing the same sample but with different modalities with a week gap to prevent potential memorization if the same worker happens to annotate the exactly identical data sample.
We estimate the time required to complete each of our HITs to be 10-15 minutes, and adjust our pay rate accordingly to \$2 or \$3 USD depending on the length of the task.
This roughly equates to a \$12 to \$15 USD per hour wage, which is above the local minimum wage for the workers.
In total we receive annotated HITs from around 80 workers for WikiHow, and 14 workers for RecipeQA.

In order to ensure annotation quality and filter potential MTurk spammers, we design a few sets to be our \textit{qualification rounds} for later on worker pool selection.
The Pearson correlation between the performance of the qualification samples and the overall HIT performance is 0.6 with p-value~\textless~0.05.
Since it is positive correlated and significant, we censor assignments with substantially low overall performance (\textless 20\% on accuracy metric), and relaunch the HITs containing those samples for a few more rounds for higher quality annotations.

Finally, since the agreement is sufficiently high (see~\secref{sec:human_annots}), we simply compute the human performance using all of the collected annotated orders from all the participated workers, which result in reasonably high human performance upper bound for our proposed sequencing task.

\subsubsection{Annotating Alternative Orders}

We deliberately ask a different set of MTurk workers than those participated in the standard performance benchmark round for annotating the alternative orders.
In total we receive HITs from around 70 workers for WikiHow, and 40 workers for RecipeQA.
The monetary rewards and other general settings follow the same procedure as in the standard performance collection.
We compute pairwise IAAs for each worker against every other workers, using the method described in~\appendixref{a-sec:iaa}, and then we place a threshold to filter out workers that tend to have too low IAAs (which is a likely indicator that a worker is either a spammer or not understanding our task well).
As the final IAAs among the selected pool of workers are sufficiently high (see~\secref{sec:human_annots}), for each instance we perform a majority vote on the annotated alternative orders to serve as the final multi-references.

\subsection{Inter-Annotator Agreements (IAA)}
% \violet{I'll move this either to the experiment section or the appendix.}
\label{a-sec:iaa}

\subsubsection{Standard Performance}

As orders concern not only positioning of the items but also more complicated relative information among the items in a sequence, we propose to measure the agreements among orders centering around the concept of \textbf{pairwise relationship}. Specifically, we transform an integer sequence order to an one-hot encoded representation of the ${N \choose 2}$ pairs of relative relations. Consider an example: suppose three items (\texttt{1}, \texttt{2}, \texttt{3}) are to be ordered, and all the pairwise relations are \{\texttt{12}, \texttt{13}, \texttt{21}, \texttt{23}, \texttt{31}, \texttt{32}\}. The transformed one-hot representation is defined as: $R_{123}$ = \{\texttt{12}: 1, \texttt{13}: 1, \texttt{21}: 0, \texttt{23}: 1, \texttt{31}: 0, \texttt{32}: 0\} = \{110100\}, \ie, $R(ij) = 1\ \text{iff} \ ij\ \text{is a valid relatively ordered pair}$. Similarly, $R_{231}$ = \{001110\}.

Using the aforementioned definition of $R$, we can compute Cohen's Kappa inter-annotator agreement score for a pair of annotated order per each instance. The overall scores can be computed by firstly taking the average of pairwise Kappa scores of annotations for each instance, and then taking the average across the entire dataset.
% We report the IAAs in~\secref{sec:benchmark_results}.

% \vspace{.3em}
% \telins{Probably will just move everything to the appendix.}

\subsubsection{Alternative Orders}

To evaluate the agreements for the alternative orders, we focus on the \textit{differences} between an order and the ground truth in their transformed representations. We first compute the \textit{one-hot difference} between an alternative order to the ground truth order, \eg suppose ground truth order is simply $o_g=$\texttt{123}, and an alternative order is $o_1=$\texttt{132}, then $R^{diff}_{o_g, o_1}$ = $abs|$\{110100\} - \{110001\}$|$ = \{000101\}. To focus on the agreements of \textit{the differences to the original ground truth}, we apply the Kappa score on a pair of orders by retaining the union of the positions where each order differ from the ground truth in their one-hot representations. For example, if $o_2=$\texttt{213}, then $R^{diff}_{o_g, o_2}$ = $abs|$\{110100\} - \{011100\}$|$ = \{101000\}, and hence the differences to the ground truth are at positions $4, 6$ from $o_1$ and $1, 3$ from $o_2$, \ie the union is $\{1, 3, 4, 6\}$. Computing the Kappa scores on $R^{diff}_{o_g, o_1}$ and $R^{diff}_{o_g, o_2}$ at these positions leads to computing the scores on lists \{0011\} and \{0110\}.

To compute the agreements of two series of alternative orders from two annotators (the series can have different lengths), we first iteratively find all the best matching pair of orders from the two series (each order in a series can only be matched \textbf{once}).
When one series contain more orders than the other, the remaining unmatched orders will be compared to the ground truth to serve as the penalty.
For a particular instance, we take the mean of all the Kappa scores (the best-matching-pair and penalty scores) as the IAA for the two annotators, as detailed in Algorithm~\ref{algo:alt_single_iaa}.
The overall IAA is computed similarly to the standard case.
% which we report in~\secref{sec:eval_alternative_orders}, with details in~\appendixref{a-sec:human}.

\newcommand{\Argmax}[1]{\underset{#1}{\operatorname{arg}\,\operatorname{max}}\;}
\begin{algorithm}[t]
\caption{Alternative Order IAA Per Instance}\label{algo:alt_single_iaa}
\begin{algorithmic}[1]
\REQUIRE $\{A_n\}_{n=1}^N$: A list of annotation series, where $A_n = \{a_{n,k}\}_{k=1}^{K_n}$ denotes $K_n$ orders annotated by $n$th worker for an instance.
\REQUIRE $f(x, y)$: IAA scoring function.
\STATE Initialize $S$: empty score list
\FOR{$i$ = 1 to N}
\FOR{$j$ = $i+1$ to N}
\STATE One-hot encode $\{a_{i,k}\}$, and $\{a_{j,k}\}$
\STATE Assume $K_i$ < $K_j$\ \ \ \ \ \textit{// otherwise swap}
\WHILE {$\{a_{i,k}\}$ not empty}
\STATE Find best match according to $R^{diff}$
\STATE $\hat{m}, \hat{n} = \Argmax{m,n}f(R^{diff}_{o_g, o_{i, m}}, R^{diff}_{o_g, o_{j, n}})$
\STATE $\{a_{i,k}\}$.pop($\hat{m}$); $\{a_{j,k}\}$.pop($\hat{n}$)
\STATE $S = S\ \cup$ score
\ENDWHILE
\WHILE {$\{a_{j,k}\}$ not empty}
\STATE $S = S\ \cup\ f(o_g, o_{j, m})$; $\{a_{j,k}\}$.pop($m$)
\ENDWHILE
\ENDFOR
\ENDFOR
\RETURN mean($S$)
\end{algorithmic}
\end{algorithm}

% \mypar{Annotation Statistics.}
% \tbref{tab:alternative_stats} lists the essential statistics of the multi-reference subsets.

\subsection{Additional Statistics}
Apart from the main sequencing task, we also ask the annotators for their confidence of predictions and if multimodality is helpful for deciding the order in the standard benchmark round.
We hereby provide two more statistics obtained from the workers: the percentages of confidence levels and which modality (modalities) helps for deciding the order.

\vspace{.3em}

\mypar{Modality Helps.}
As which modality is potentially more helpful, we include the percentages of each answer category in~\tbref{tab:modality_help}.
It can be noticed that majority of workers (\textgreater~60\%) think that multimodal (both modalities) is helpful, and especially in the recipe data, there are \textgreater~90\% of workers indicating the effectiveness of utilizing multimodal inputs.

\vspace{.3em}

\mypar{Confidence Levels.}
As shown in~\tbref{tab:confidence_level}, majority of workers feel at least fairly confident (score of 4) about their predictions, which can justify the validity of our selection of golden-test-set.

\begin{table}[t!]
\centering
\small
\scalebox{.95}{
    \begin{tabular}{c|cccc}
    \toprule
    \textbf{Dataset} & Both & Text-Only & Image-Only & Neither \\
    \midrule % \\[-1.1em]
    RecipeQA & 90.4 & 1.0   & 8.6 & 0.0 \\
    WikiHow  & 62.9  & 33.7 & 2.4 & 1.0 \\
    \bottomrule
    \end{tabular}
}
\caption{\footnotesize
\textbf{Which modality helps?}
We compute the percentage of each answer category.
In both datasets, majority of the annotations indicate that both modality are helpful for deciding the orders. 
}
\label{tab:modality_help}
\end{table}

\begin{table}[t!]
\centering
\small
\scalebox{1.0}{
    \begin{tabular}{l|rr}
    \toprule
    \textbf{Confidence Level} & \textbf{WikiHow} & \textbf{RecipeQA} \\
    \midrule % \\[-1.1em]
    5 (Very)       & 54.61  & 64.75 \\
    4 (Fairly)     & 27.38  & 23.00 \\
    3 (Moderately) & 12.24  &  7.00 \\
    2 (Somewhat)   &  5.21  &  4.75 \\
    1 (Not-At-All) &  0.56  &  0.50 \\
    \bottomrule
    \end{tabular}
}
\caption{\footnotesize
\textbf{Confidence Level Statistics (\%):} In both datasets, majority (\textgreater~80\%) of the annotators indicate at least~\textgreater~4 (fairly) confidence level, which can help justify the validity of the human performance.
}
\label{tab:confidence_level}
\end{table}

\section{Additional Results}
\label{a-sec:add_res}

\subsection{Qualitative Inspections}
\label{a-sec:quali}

\begin{figure*}[t!]

\begin{subtable}{\textwidth}
\centering
\centering
    \includegraphics[width=.95\columnwidth]{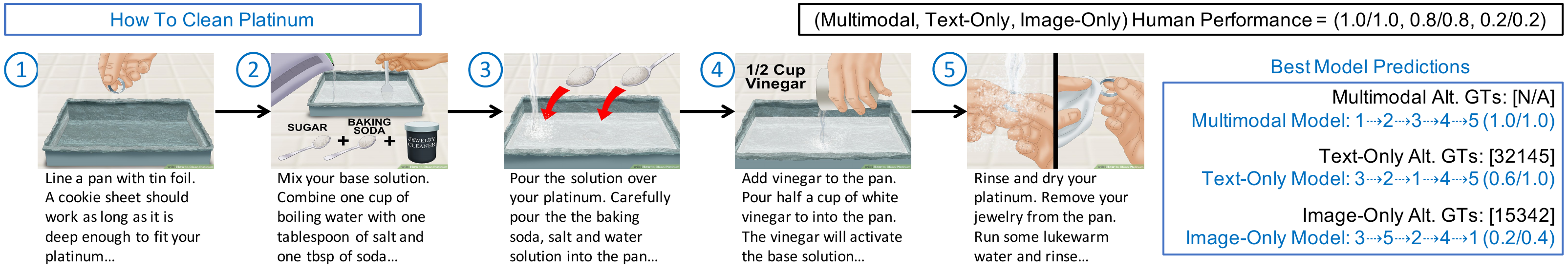}
\caption{\footnotesize Home and Garden Sample}
\label{fig:quali_main_1}
\end{subtable}

\begin{subtable}{\textwidth}
\centering
\centering
    \includegraphics[width=.95\columnwidth]{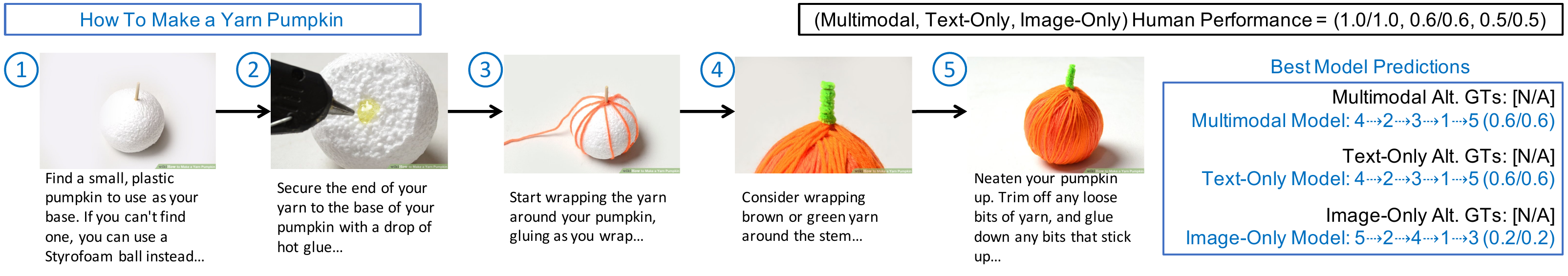}
\caption{\footnotesize Hobbies and Crafts Sample}
\label{fig:quali_main_2}
\end{subtable}

\begin{subtable}{\textwidth}
\centering
\centering
    \includegraphics[width=.95\columnwidth]{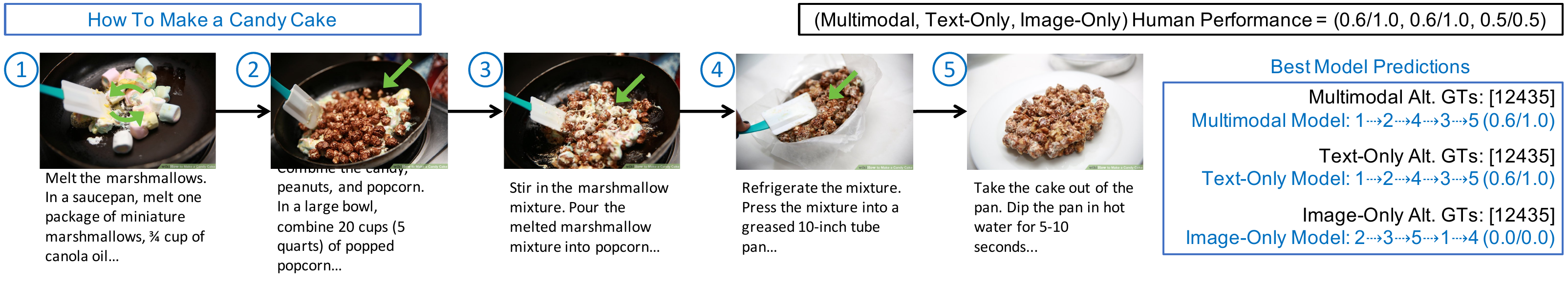}
\caption{\footnotesize Recipe Sample}
\label{fig:quali_main_3}
\end{subtable}

\caption{\footnotesize
    \textbf{Qualitative examples:} We show some qualitative samples of our dataset associated with human and model predictions, and the annotated multi-reference ground truths. The texts are truncated to fit into the box shown in each sample. The performance are: (single-reference, multi-reference) \textbf{accuracy} metric respectively.
}
\label{fig:quali_main}
\end{figure*}

\figref{fig:quali_main} shows a few qualitative examples in different categories.~\figref{fig:quali_main_1} shows that while step 1 and 3 may seem confusing if only looking at the texts, the images can help deciding the proper order, whereas models may fail to grasp such multimodal information in~\figref{fig:quali_main_2}.
In~\figref{fig:quali_main_3} we show an example where multi-reference benefits both humans and the models, although in reality it should be more commonsensical to \textit{stir} before \textit{refrigerating} the mixtures.

\subsection{Image-Only Multi-References}

We also provide the detailed multi-reference performance break down on the image-only modality using the best performing models in~\tbref{tab:model_performances}, CLIP, in~\tbref{tab:multiref_perf_img_only} for references.

\begin{table*}[th!]
\centering
\small
\scalebox{.89}{
    \begin{tabular}{cr|cccccc|cccccc}
    \toprule
    \multicolumn{1}{c}{\multirow{3}{*}{\textbf{Modality}}}
    & \multicolumn{1}{c|}{\multirow{3}{*}{\textbf{Subset}}}
    & \multicolumn{6}{c|}{\textbf{WikiHow Golden-Test-Set (Size: 300)}}
    & \multicolumn{6}{c}{\textbf{RecipeQA Golden-Test-Set (Size: 100)}} \\
    &
    & \multicolumn{2}{c}{Acc$\uparrow$} & \multicolumn{2}{c}{PMR$\uparrow$} & \multicolumn{2}{c|}{L$_{r}\uparrow$}
    & \multicolumn{2}{c}{Acc$\uparrow$} & \multicolumn{2}{c}{PMR$\uparrow$} & \multicolumn{2}{c}{L$_{r}\uparrow$} \\
    &
    & single & multi & single & multi & single & multi
    & single & multi & single & multi & single & multi \\
    
    \midrule

    \multirow{6}{*}{Image-Only}
    & \multirow{1}{*}{Single}
      & 28.38 & --- & 5.07 & --- & 1.97 & --- & 49.89 & --- & 17.24 & --- & 2.47 & --- \\
    & \multirow{1}{*}{Multi.}
      & 26.67 & 39.17 & 4.17 & 8.33 & 1.83 & 1.92 & 29.23 & 40.00 & 7.69 & 7.69 & 1.92 & 2.31 \\
    & \multirow{1}{*}{All}
      & 28.24 & 29.24 & 5.00 & 5.33 & 1.96 & 1.97 & 47.2 & 48.60 & 16.00 & 16.00 & 2.40 & 2.45 \\
    \cline{2-14} \\[-.8em]
    & \multirow{1}{*}{Single$\dagger$}
      & 68.47 & --- & 48.36 & --- & 3.54 & --- & 81.61 & --- & 66.67 & --- & 4.10 & --- \\
    & \multirow{1}{*}{Multi.$\dagger$}
      & 64.58 & 75.83 & 37.50 & 56.25 & 3.19 & 3.71 & 72.31 & 79.23 & 50.00 & 61.54 & 3.50 & 3.88 \\
    & \multirow{1}{*}{All$\dagger$}
      & 68.16 & 69.06 & 47.49 & 48.99 & 3.51 & 3.55 & 80.40 & 81.30 & 64.50 & 66.00 & 4.02 & 4.07 \\

    \bottomrule
    \end{tabular}
}
\footnotesize{\\$\ast$ The size of the \textbf{Multi.} subsets are: 24/300 in WikiHow and 13/100 in RecipeQA.}
\caption{
\footnotesize
\textbf{Multi-reference performance on image-only modality:} $\dagger$ denotes human performance. The denotations are same as the~\tbref{tab:multiref_perf}.
Results are reported using the best performing image-only models from~\tbref{tab:model_performances}.
}
\label{tab:multiref_perf_img_only}
% \vspace{-1.5em}
\end{table*}

\section{More Model Details}
\label{a-sec:models}

\mypar{Multimodal Model Considerations.}
\citet{bugliarello-etal-2020-multimodal} suggests that many V\&L models can achieve similar downstream performance if well trained, and thus we consider the models presented in this work, VisualBERT and CLIP-ViL, due to their simplicity of adapting to our sequencing task, as well as their main differences being how the visual inputs are encoded (via standard object detector networks or patch-based models like CLIP), which suits our proposed objectives well.

\vspace{.3em}

\mypar{Swapping-Based Predictions.}
In~\secref{sec:isp} we mention that we do not observe necessary improvements when swapping the textual contents. Our hypothesis is that the pairwise loss function applied in the BERSON module already takes care of this especially for the textual contents. And that the stronger discourse-level hints inherent in the textual descriptions may make this operation unnecessary. On the other hand, both image and multimodal alignment does not share this similar property with the texts, and hence this reasons why swapping the visual modality suffices this particularly pretraining objective.

\subsection{Training \& Implementation Details}
\label{a-sec:impl}

\begin{table*}[t]
\centering
\footnotesize
\scalebox{0.9}{
    \begin{tabular}{clccccccc}
    \toprule
    \multirow{2}{*}{\textbf{Modalities}} & \multicolumn{1}{c}{\multirow{2}{*}{\textbf{Models}}} & \multirow{2}{*}{\textbf{Batch Size}} & \multirow{2}{*}{\textbf{Initial LR}} & \multirow{2}{*}{\textbf{\# Training Epochs}} & \textbf{Gradient Accu-} & \multirow{2}{*}{\textbf{\# Params}}  \\
    & & & & & \textbf{mulation Steps} & \\
    \midrule
    \multirow{2}{*}{Image-Only}
    & ResNet & 4 & $5 \times 10^{-6}$ & 5  & 1 & 112.98M \\
    & CLIP   & 4 & $5 \times 10^{-6}$ & 5  & 1 & 88.08M \\
    \midrule
    \multirow{1}{*}{Text-Only}
    & RoBERTa & 4 & $5 \times 10^{-6}$ & 5  & 1 & 393.16M \\
    \hline \\[-1em]
    \multirow{2}{*}{Multimodal}
    & VisualBERT      & 4 & $5 \times 10^{-6}$ & 10 & 1 & 421.32M \\
    & CLIP-ViL & 4 & $5 \times 10^{-6}$ & 10 & 1 & 497.40M \\
    \hline \\[-1em]
    \multirow{1}{*}{Image-Only Pretrain}
    & CLIP   & 4 & $1 \times 10^{-5}$ & 5 & 1 & 68.09M \\
    \hline \\[-1em]
    \multirow{1}{*}{Text-Only Pretrain}
    & RoBERTa    & 4 & $1 \times 10^{-5}$ & 5 & 1 & 355.36M \\
    \hline \\[-1em]
    \multirow{2}{*}{Multimodal Pretrain}
    & VisualBERT      & 4 & $1 \times 10^{-5}$ & 5 & 1 & 383.52M \\
    & CLIP-ViL & 4 & $1 \times 10^{-5}$ & 5 & 1 & 465.50M \\
    \bottomrule
    \end{tabular}
}
\caption{\footnotesize
\textbf{Hyperparameters in this work:} \textit{Initial LR} denotes the initial learning rate. All the models are trained with Adam optimizers~\cite{kingma2014adam}. We include number of learnable parameters of each model in the column of \textit{\# params}.
}
\label{tab:hyparams}
\end{table*}

\begin{table*}[t]
\centering
\footnotesize
\begin{tabular}{ccccc}
    \toprule
    \textbf{Type} & \textbf{Batch Size} & \textbf{Initial LR} & \textbf{\# Training Epochs} & \textbf{Gradient Accumulation Steps} \\
    \midrule
    \textbf{Bound (lower--upper)} & 2--8 & $1 \times 10^{-5}$--$1 \times 10^{-6}$ & 3--10 & 1--2 \\
    \midrule
    \textbf{Number of Trials} & 2--4 & 2--3 & 2--4 & 1--2 \\
    \bottomrule
\end{tabular}
\caption{\footnotesize
\textbf{Search bounds} for the hyperparameters of all the models.
}
\label{tab:search}
\end{table*}

\mypar{Training Details.}
All the models in this work are trained on a single Nvidia A100 GPU\footnote{https://www.nvidia.com/en-us/data-center/a100/} on a Ubuntu 20.04.2 operating system.
The hyperparameters for each model are manually tuned against different datasets, and the checkpoints used for testing are selected by the best performing ones on the held-out development set, which is constructed using the method described in~\appendixref{a-sec:splits}.

\vspace{.3em}

\mypar{Implementation Details.}
The implementations of the transformer-based models are extended from the HuggingFace\footnote{https://github.com/huggingface/transformers}~code base~\cite{wolf-etal-2020-transformers}, and our entire code-base is implemented in PyTorch.\footnote{https://pytorch.org/}
The computer vision detector model used in one of our image-only encoders, ResNet-based Faster-RCNN~\cite{ren2016faster}, adopts the detectron2 open sourced module, and their pretrained weights are obtained from the official implementation from Facebook AI Research.\footnote{https://github.com/facebookresearch/detectron2}
Implementation of BERSON modules are adapted from the original author's implementation, where more details can be found in their paper.
Implementation of the VisualBERT is obtained from the \textit{MMF}\footnote{https://github.com/facebookresearch/mmf} framework from Facebook AI Research, and CLIP-ViL model is obtained and adapted from the original author's released code repository.\footnote{https://github.com/clip-vil/CLIP-ViL}
We use this same repository for the image-only encoder CLIP.

\subsection{Hyperparameters}

For the sequencing task, we train all the models for 5 or 10 (for multimodal models) epochs for all the model variants, where the training time varies from 2-4 hours for the text-only models and 6-8 hours for the multimodal models.
We list all the hyperparameters used in~\tbref{tab:hyparams}. We also include the search bounds and number of trials in~\tbref{tab:search}, that all of our models adopt the same search bounds and ranges of trials.

\subsection{WikiHow Images}
\label{a-sec:img_feat}

\begin{figure*}[h!]

\begin{subtable}{.48\textwidth}
\centering
\centering
    \includegraphics[width=.8\columnwidth]{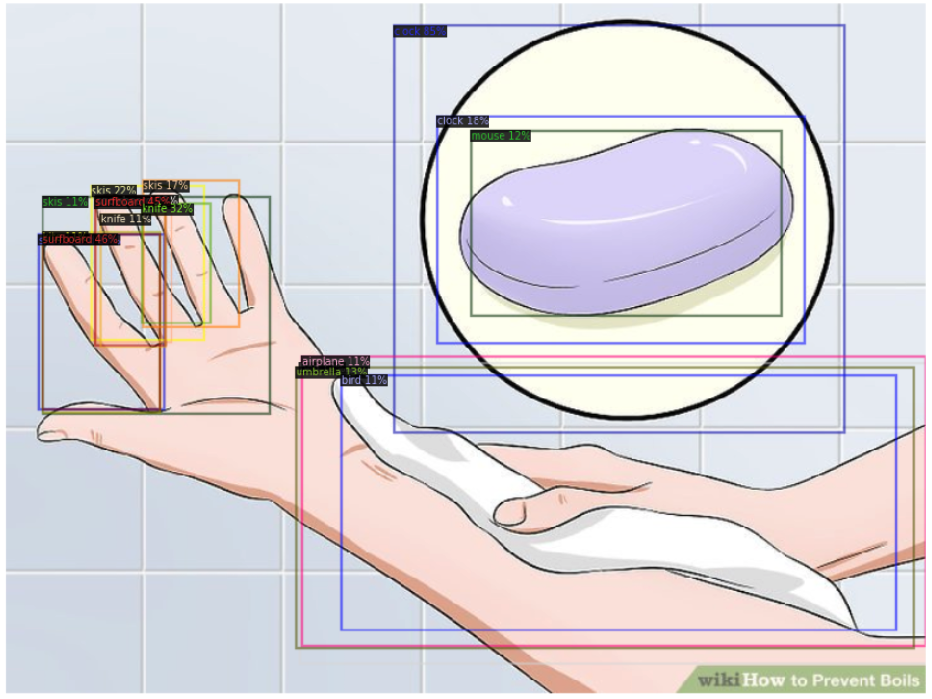}
\caption{\footnotesize Detected Image Regions 1}
\label{fig:detectron2_1}
\end{subtable}
\quad
\begin{subtable}{.48\textwidth}
\centering
\centering
    \includegraphics[width=.8\columnwidth]{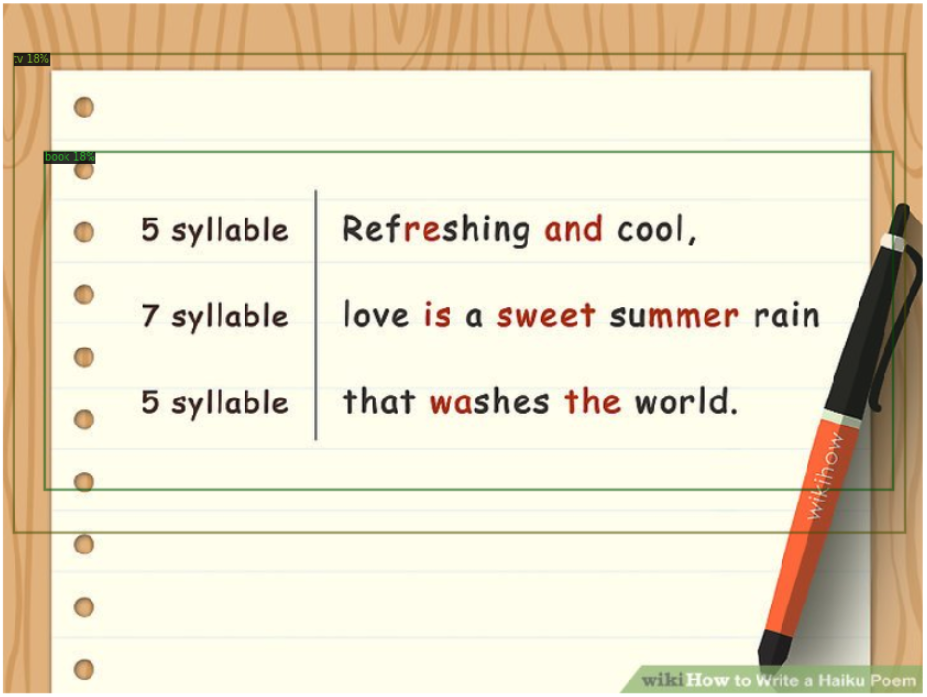}
\caption{\footnotesize Detected Image Regions 2}
\label{fig:detectron2_2}
\end{subtable}

\caption{ \footnotesize
    \textbf{Proposed image regions by Detectron2:} We show some examples that even these synthetic and cartoon-ish images in the WikiHow dataset can provide meaningful representations which can be utilized by strong pretrained object detection modules. We show few top-detected objects with their bounding boxes and predicted classes. Note that while the classes may be wrongly predicted, the proposed regions are all meaningful.
}
\label{fig:detectron2}
\end{figure*}

Although the images in WikiHow can often be synthetic or "cartoon-ish", we observe that modern object detectors can still propose meaningful regions, regardless of whether the object class prediction is sensible or not.
We include some predicted bounding boxes in~\figref{fig:detectron2} for references.
And hence, although there may be concerns on suboptimal visual understanding from these images, we do believe both of our ResNet and CLIP visual encoders can extract reasonably useful features.

\section{Releases \& Codes}
\label{a-sec:release}
The scraped WikiHow dataset will be released upon acceptance, along with a clearly stated documentation for usages. We will also release the code for processing the RecipeQA dataset particularly for our procedure sequencing task, where the original dataset can be obtained from their project website.\footnote{https://hucvl.github.io/recipeqa/}
If permitted by the authors of the BERSON model, we will also release the cleaned code repository which encompasses the majority of the implementations in this work upon acceptance.
% \footnote{The code repository included in the appendix is tentatively a minimum functional version.}.
We hope that by sharing the datasets and their essential tools, more interest could be drawn into research on multimodal procedure understanding and its future research directions.

\end{document}